\documentclass[times, review, 10pt]{elsarticle}

\usepackage{amssymb}
\usepackage{amsmath}
\usepackage{wrapfig}

\usepackage{multirow} 
\usepackage[table]{xcolor}
\usepackage{amsmath}
\usepackage[utf8]{inputenc} 
\usepackage[T1]{fontenc}    
\usepackage{hyperref}       
\usepackage{url}            
\usepackage{booktabs}       
\usepackage{amsfonts}       
\usepackage{nicefrac}       
\usepackage{microtype}      
\usepackage{xcolor}         
\usepackage{upgreek}
\usepackage{graphicx}
\usepackage{wrapfig}
\usepackage{lipsum}  
\usepackage{caption}
\usepackage{blindtext}
\usepackage{tabularray}
\usepackage{array}
\usepackage{pifont}
\usepackage{colortbl}
\definecolor{lightblue}{RGB}{16, 113, 188}  
\usepackage{CJKutf8}
\usepackage{algorithm}
\usepackage{algorithmic}
\usepackage[table]{xcolor}
\usepackage{graphicx}
\definecolor{lightb}{RGB}{225,240,255}
\definecolor{darkb}{RGB}{190,220,255}
\usepackage{longtable}
\usepackage{caption}
\usepackage{needspace}
\journal{Pattern Recognition}

\begin{document}





\begin{frontmatter}

\title{\fontsize{14pt}{16.8pt}\selectfont Sigma: Semantically Informative Pre-training for Skeleton-based Sign Language Understanding}

\author[label1]{\footnotesize Muxin Pu\corref{cor1}}
\ead{muxin.pu@monash.edu}

\author[label1]{\footnotesize Mei Kuan Lim}
\ead{lim.meikuan@monash.edu}

\author[label1]{\footnotesize Chun Yong Chong}
\ead{chong.chunyong@monash.edu}

\author[label2]{\footnotesize Chen Change Loy}
\ead{ccloy@ntu.edu.sg}

\cortext[cor1]{\footnotesize Corresponding author.}

\affiliation[label1]{
  organization={School of Information Technology, Monash University},
  addressline={Jalan Lagoon Selatan},
  city={Subang Jaya},
  postcode={47500},
  state={Selangor},
  country={Malaysia}
}

\affiliation[label2]{
  organization={S-Lab, Nanyang Technological University},
  addressline={Nanyang Avenue},
  city={Singapore},
  postcode={639798},
  country={Singapore}
}





\begin{abstract}
Pre-training has proven effective for learning transferable features in sign language understanding (SLU) tasks. Recently, skeleton-based methods have gained increasing attention because they can robustly handle variations in subjects and backgrounds without being affected by appearance or environmental factors. Current SLU methods continue to face three key limitations: 1) weak semantic grounding, as models often capture low-level motion patterns from skeletal data but struggle to relate them to linguistic meaning; 2) imbalance between local details and global context, with models either focusing too narrowly on fine-grained cues or overlooking them for broader context; and 3) inefficient cross-modal learning, as constructing semantically aligned representations across modalities remains difficult. To address these, we propose Sigma, a unified skeleton-based SLU framework featuring: 1) a sign-aware early fusion mechanism that facilitates deep interaction between visual and textual modalities, enriching visual features with linguistic context; 2) a hierarchical alignment learning strategy that jointly maximises agreements across different levels of paired features from different modalities, effectively capturing both fine-grained details and high-level semantic relationships; and 3) a unified pre-training framework that combines contrastive learning, text matching and language modelling to promote semantic consistency and generalisation. \textbf{Sigma} achieves new state-of-the-art results on isolated sign language recognition, continuous sign language recognition, and gloss-free sign language translation on multiple benchmarks spanning different sign and spoken languages, demonstrating the impact of semantically informative pre-training and the effectiveness of skeletal data as a stand-alone solution for SLU.
\end{abstract}

\begin{highlights}
\item We present a unified framework for sign language understanding.
\item We introduce a semantically informative pre-training for sign language understanding.
\item We enhance cross-modal representations by improving visual-textual alignment.
\end{highlights}


\end{frontmatter}



\section{Introduction} \label{sec:introduction}
Sign languages are the primary means of communication for around 70 million people with hearing or speech impairments, spanning more than 200 SLs worldwide \citep{WHO, WFD}. SLs remain challenging for the general public to master due to the global diversity and complex structure, which encompasses rapid and intricate hand gestures, body postures, as well as facial expressions. The ultimate goal of sign language understanding (\textbf{SLU}) is to comprehend SLs at the levels of words, phrases, as well as sentences anytime and anywhere for the impaired community, enabling barrier-free communication for them. Achieving this goal requires the development of models capable of interpreting these visual signals in alignment with the unique linguistic structure of SLs. SLU typically comprises three core tasks: isolated sign language recognition (\textbf{ISLR}), which recognises sign glosses \footnote{A sign gloss is a textual label that represents the meaning of a sign sequence using a word or phrase.} \citep{hu2021signbert, hu2023signbertplus, pu2024siformer}; continuous sign language recognition (\textbf{CSLR}), which aligns unsegmented sign sequences with sign glosses \citep{hu2021signbert, Zuo_2022_CVPR, fu2025improving}; and sign language translation (\textbf{SLT}), which converts sign sequences into sentences \citep{zhou2021improving, zhou2023gloss, fu2025improving}. These tasks demand both fine-grained visual recognition and strong contextual understanding. 

Recently, SLU research has progressively shifted from fully supervised learning toward the development of effective pre-training paradigms, commonly referred to as sign language pre-training (\textbf{SLPT}) \citep{zhou2023gloss, hu2023signbertplus, zhou2024scaling, Li2025sign, fu2025improving}. These methods present a promising direction by enabling models to learn transferable representations directly from sign language data, thereby significantly reducing the reliance on manual annotations, such as gloss annotations, temporal boundaries or clip-level supervision. By capturing structural and temporal regularities during the pre-training stage, models gain generalizable knowledge that accelerates convergence and enhances performance on a wide range of downstream SLU tasks. Consequently, SLPT serves as a foundational step toward building unified and scalable SLU frameworks. Despite their potential, current SLP-based SLU methods continue to face significant limitations.

First, the \textbf{lack of semantic grounding} in visual representations remains a major challenge in advancing SLU. While dense geometric features in skeletal data, such as hand trajectories, body movements, and facial expressions, provide important visual cues, they often carry limited linguistic meaning. Most existing skeleton-based SLU methods focus on capturing these low-level patterns from skeletal data, treating sign language primarily as a visual signal and paying little attention to the underlying linguistic structure \citep{hu2021signbert, hu2023signbertplus, zhao2024masa, pu2024siformer}. Although such models may capture low-level motion patterns, they struggle to model the relationship between these geometric features and their intended semantic roles. This disconnect weakens the ability of models to produce accurate and meaningful interpretations. Addressing this issue requires enriching visual features with semantic grounding, allowing the model to understand both the appearance and the purpose of each gesture. Doing so helps bridge the gap between visual representation and language understanding, making the model capable of supporting accurate recognition and fluent translation.


Second, the \textbf{imbalance between local-global feature modelling} remains a persistent challenge in SLU, which inherently spans both recognition and translation tasks. Accurately distinguishing subtle variations in sign language gestures requires capturing fine-grained local motion patterns, while achieving coherent understanding necessitates preserving high-level global semantics. Balancing these two levels of representation is inherently difficult but critical \citep{liu2013global}. Global semantic modelling plays a key role in resolving ambiguities between visually similar sign language gestures, particularly in continuous streams where the boundaries of sign glosses are unclear and context determines meaning. In such cases, local features alone are inadequate. Conversely, precise local detail extraction is equally vital, as small variations in hand gestures, body postures, or facial expression can dramatically alter meaning and grammatical structure. Even minor changes in motion intensity may shift interpretation and degrade translation quality \citep{camgoz2018neural}. Therefore, robust SLU demands a mechanism that jointly models both local and global features in a balanced manner. 

Third, \textbf{inefficient cross-modal representation learning} remains a critical bottleneck for advancing SLU. Compared to traditional video understanding, SLU from RGB videos is more challenging because gestures and facial expressions are more intricate or rapid than general human actions or scene changes \citep{hu2021signbert, hu2023signbertplus}. Constructing structured, semantically aligned representations from raw visual streams is difficult, as models are easily distracted by background details or appearance variations rather than focusing on the linguistic cues that carry meaning \citep{hu2021signbert, hu2023signbertplus, pu2024siformer}. This inefficiency weakens the alignment between dynamic gestures and textual semantics while imposing heavy computational and storage costs, ultimately limiting the scalability of SLU and slowing progress in sign language production and generation. Skeletal data offers a promising alternative to RGB videos \citep{hu2021signbert, hu2023signbertplus, pu2024siformer}. It intentionally prioritises the essential spatial-temporal dynamics of SL, which are the core semantic carriers in SLU. Modest estimation variances in skeletal data can improve generalisation across diverse real-world motion patterns, and by abstracting away visual noise such as lighting, background clutter, and appearance biases, skeletal representations provide cleaner, more relevant inputs with stronger privacy guarantees.

Collectively, there is a need for an approach that enhances meaningful semantic grounding, promotes balanced feature modelling, and supports effective cross-modal representation learning for skeleton-based SLU. Visual illustrations of our motivation are provided in the Appendix. To overcome these limitations, this paper proposes the following solutions:
\begin{itemize}
    \item We introduce \textbf{a sign-aware early fusion mechanism} that enables bidirectional interaction between visual and textual features during the encoding stage. This encourages the model to learn semantically enriched visual representations, improving modality alignment and deepening contextual comprehension.
    \item We propose \textbf{a hierarchical alignment learning strategy}, which learns representations by maximising agreement across modalities. This enables the model to capture fine-grained visual cues and high-level semantic structures, supporting accurate recognition and fluent translation.
	\item We design \textbf{a skeleton-based unified cross-modal pre-training framework} that facilitates efficient and flexible representation learning across multiple tasks. By jointly optimising contrastive learning, text matching, and language modelling within a shared space, the framework improves semantic alignment as well as boosts transferability and generalisation across diverse downstream SLU tasks.
\end{itemize}

\section{Related work}
\subsection{Semantically informative visual feature} 
Learning semantically informative visual features from sign language sequences is crucial for understanding. This is particularly important in resolving the representation density problem, where visually similar sign language gestures, differing only slightly in motion or expression, tend to cluster closely in feature space \citep{ye2024improving}. Incorporating linguistic and contextual cues into visual representations helps mitigate feature overlap and enables the model to learn more separable and discriminative features. This can lead to improved performance in both recognition and translation tasks, especially in cases where subtle visual differences correspond to distinct meanings. Prior works such as TSPNet \citep{li2020tspnet}, GLE-Net \citep{hu2021global}, HST-GNN \citep{kan2022sign} and SignCL \citep{ye2024improving} have made progress in temporal modelling, global context extraction, and multi-perspective graph-based reasoning. However, learning and embedding semantically rich visual features in a way that generalises across tasks remains an open challenge in advancing SLU.

\subsection{Sign language understanding} 
SLU has been widely studied through task-specific methods. Prior works for ISLR have applied spatial-temporal modelling to improve accuracy \citep{hu2021hand, li2020transferring, zuo2023natural}. Recent models for CSLR address co-articulation and gloss boundary ambiguity using CTC-based or sequence-to-sequence frameworks \citep{min2021visual, hu2023adabrowse, jiao2023cosign}. For SLT, gloss-based approaches rely on intermediate gloss annotations \citep{camgoz2020sign, zhou2021spatial}, while emerging gloss-free methods adopt pre-training and large language models to reduce annotation requirements and improve generalisation \citep{zhou2023gloss, wong2024sign2gpt, gong2024llms}. In this study, we focus on the gloss-free SLT paradigm and aim to enhance its effectiveness by learning semantically rich visual representations aligned with textual outputs. In contrast to prior task-specific methods, we propose a unified framework capable of performing all aforementioned SLU tasks. A central design motivation lies in the differing representational needs across tasks of recognition and translation.

\subsection{Sign language pre-training} 
SLPT methods employ pretext tasks to learn useful representations from sign language data, improving downstream performance. Self-supervised models like SignBERT \citep{hu2021signbert, hu2023signbertplus} use masking and reconstruction to capture visual patterns from unlabeled videos but often lack sufficient semantic grounding. To address this, MMTLB \citep{chen2022simple} introduces multi-task training across sign-to-gloss, gloss-to-text, and sign-to-text objectives, while GFSLT-VLP \citep{zhou2023gloss} uses contrastive learning for sign-text alignment. More recent efforts, including MSLU \citep{zhou2024scaling} and C$^{2}$RL \citep{Zuo_2022_CVPR}, incorporate keypoint reconstruction and language modelling to enhance semantic representation. Despite these advances, most approaches remain task-specific, limiting scalability. Moreover, they often struggle to balance modality-specific encoding with effective cross-modal transfer, both of which are essential for developing unified and generalisable SLU systems.

\section{Method}
\begin{figure*}[htbp]
  \centering
  \includegraphics[scale=0.45]{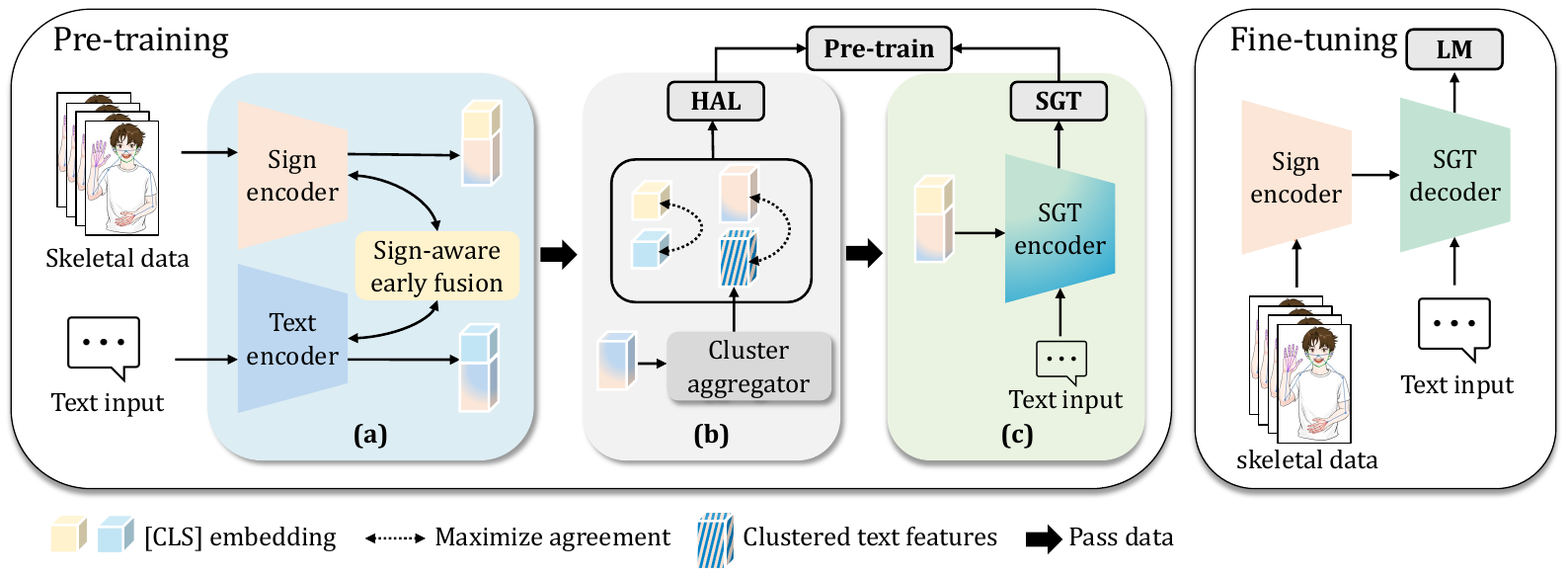}
  \caption{Overview of Sigma. (a) SignEF enhances visual-linguistic alignment by injecting cross-modal features into sign and text encoders. (b) HAL is used to maximise global and local cluster agreement. (c) SGT encoder jointly optimises sign-text matching and language modelling. During fine-tuning, both the sign and SGT encoders are reused across SLU tasks.}
  \label{fig:pipeline}
\end{figure*}

Sigma consists of two stages: pre-training and fine-tuning (Figure \ref{fig:pipeline}). In pre-training, we introduce sign-aware early fusion (\textbf{SignEF}) for deep bidirectional cross-modal interaction, hierarchical alignment learning (\textbf{HAL}) for multi-level semantic alignment to capture both coarse and fine-grained semantic correspondences, and a sign-grounded text (\textbf{SGT}) encoder jointly trained with text matching and language modelling to enhance semantic consistency and linguistic fluency. The sign encoder is fully transferred, and the SGT encoder is reused in the unified fine-tuning, enabling consistent and efficient adaptation across SLU tasks, including ISLR, CSLR, and SLT.

\subsection{Preliminaries}
We use paired skeletal data and their corresponding text(s) for both the pre-training and fine-tuning stages. The text(s) are tokenised before being fed into the text encoder. The skeletal data are 2D keypoints estimated from sign language videos using RTM-Pose \citep{jiang2023rtmpose}. Part-specific ST-GCNs \citep{yan2018spatial} are used to model both joint interdependencies and motion dynamics. Following these ST-GCNs, the raw skeletal input $S^{\text{raw}}_p \in \mathbb{R}^{L \times N_p \times C}$ is projected into a compact feature $S_p \in \mathbb{R}^{L \times D}$, where $L$ is the sequence length,\vspace{+0.1em} $N_p$ is the number of keypoints in a group that $p \in \{lh, rh, b, f\}$ (left hand, right hand, body, and face), $C$ is the visual input dimension, and $D$ is the projected dimension. The sign encoder input $S \in \mathbb{R}^{L \times 4D}$ is formed by concatenating features from all groups and serves as the visual input for the two-stage training of Sigma. 

\subsection{Sign language pre-training}
We initialise Sigma with pre-trained mT5 base \citep{xue2020mT5} to leverage knowledge from large-scale textual corpora for enhanced visual–linguistic alignment. All mT5 parameters are fully trained during both the pre-training and fine-tuning stages. This end-to-end optimisation allows the textual backbone to adapt to the characteristics of sign-derived inputs.
\subsubsection{Sign-aware early fusion mechanism} 

\begin{figure}[h]
    \centering
    \includegraphics[width=0.5\textwidth]{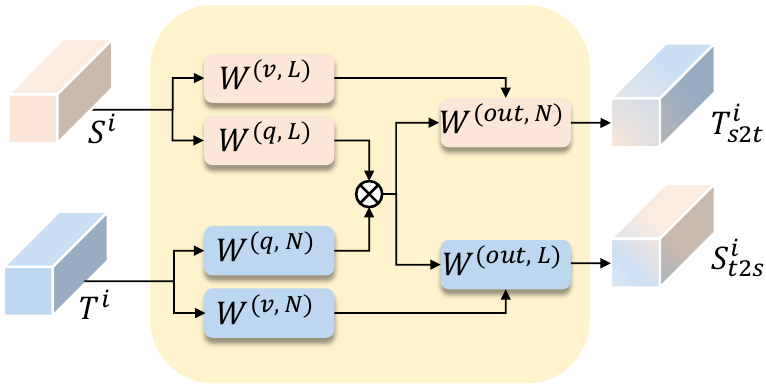}
    \caption{SignEF promotes progressive visual-linguistic interaction with parameters ${ W^{(x, L)}, W^{(x, N)} : x \in \{q, v, \text{out} \} }$, analogous to query, value, and output projections by \citep{vaswani2017attention}.}
  \label{fig:SignEF}
\end{figure}

A key challenge in skeleton-based SLU is aligning geometric gesture features with textual semantics. Inspired by \citep{vaswani2017attention, li2022glip}, we propose SignEF, which enriches sign language representations by introducing cross-modal interaction at the encoding stage, fostering more expressive and semantically aligned features. Specifically, SignEF deploys cross-attention and injects textual cues into visual encoding layers, enabling the model to perform deep and structured representation learning across modalities.

Let $S^i$ and $T^i$ denote the visual and textual features from the $i$-th layers of the sign and text encoders. Their fusion outputs, $S^{i}_{t2s}$ (text-to-sign) and $T^{i}_{s2t}$ (sign-to-text), are fed back into the encoders, with early fusion applied at the last few layers of the encoders. The process is defined as:

\begin{equation} \label{eq:SignEF}
\begin{split}
    & S^{i}_{t2s}, T^{i}_{s2t} = SignEF(S^{i}, T^{i}) \\
    & X^{i+1} = Mo\text{-}Encoder_{i+1}\left(X^{i} + X^{i}_{\bar{m}2m}\right)
\end{split}
\end{equation}


where $X$ denotes either sign ($S$) or text ($T$) features, with $Mo \in \{Sign, Text\}$ indicating the target modality and $m \in \{s, t\}$ the source. The SignEF module lets one modality attend to the other, computing cross-modal context features. As shown in Figure \ref{fig:SignEF}, attention heads share parameters across the final SGT layers. This parameter-sharing design prompts fine-grained visual–linguistic alignment while keeping the model efficient.

\subsubsection{Hierarchical alignment learning} \label{sec:hal}
\begin{figure}[htb]
  \centering
  \includegraphics[width=0.5\textwidth]{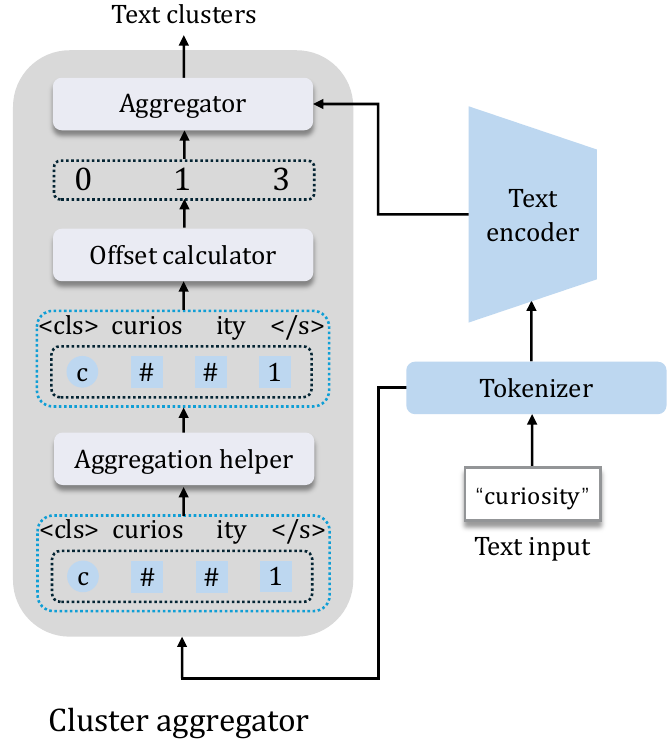}
  \caption{The overview of the cluster aggregator module. It converts sub-word token embeddings into cluster-level representations by grouping tokens, mapping them with offset indices, and aggregating hidden features for semantic alignment with visual inputs (check the Appendix for details).}
  \label{fig:cluster_aggreator}
\end{figure}

Balancing local and global feature modelling is essential for SLU, where recognition and translation require attention to both detailed and holistic semantics. Inspired by contrastive learning \citep{chen2020simple, radford2021learning, li2022blip, hou2024bagformer}, we introduce HAL as a core strategy for pre-training. HAL maximises agreement between sign-text pairs at both the global and local cluster levels. The similarity is computed as:

\begin{equation}
\mathbf{M}^{x}_{s2t} = sim(S_{f}, T_{f}; \phi) = 
\begin{cases}
    g_s(s_{cls})^\top g_t(t_{cls}), \text{ if } x = g \text{} & \text{(2.1)} \\
    \sum_{i=1}^n max_{j \in \{1, \dots, k\}} \left( g_s(s_i)^\top g_t(c_j) \right), \text{ if } x = l \text{} & \text{(2.2)}
\end{cases}
\end{equation} \label{eq:sim}

where $f$ denotes features, $l$ indicates local, and $g$ means global. Globally, we align the class-token representations $s_{cls}$ and $t_{cls}$ from the sign and text encoders. These class tokens are projected into a shared embedding space using projection heads $g_s$ and $g_t$, enabling the model to capture coarse-grained semantic relationships across modalities. A cluster aggregator (see Figure \ref{fig:cluster_aggreator}) compresses text tokens into $k$ clusters, which serve as compact semantic units to correspond to the $n$ sign tokens. To focus on cross-modality alignment, we compute the maximum similarity between each sign token $s_i$ and all text clusters $c_j$. Locally, HAL enhances fine-grained interaction by computing local cluster-wise similarity between sign features and clustered text features. 

\begin{algorithm}[thp]
\caption{Cluster-wise sign-to-text similarity (check Figure \ref{fig:local-cw-sim} for visualisation)}
\label{alg:cluster_sim}
\fontsize{10pt}{10pt}\selectfont 
\begin{algorithmic}[1]
\STATE \textbf{Input:} Sign tokens  $\{S_b \in \mathbb{R}^{N\times D}\}_{b=1}^B$,
       Textual clusters $C\!\in\!\mathbb{R}^{B\times K\times D}$
\STATE \textbf{Output:} Cluster-wise sign-to-text similarity $\mathbf{M}^{l}_{\mathrm{s2t}}$

\STATE Initialise $\mathbf{M}^{l}_{\mathrm{s2t}}\leftarrow \mathbf{0}^{B\times B}$
\FOR{$i=1$ \TO $B$}
  \STATE $M\leftarrow S_b\,C^\top$ \hfill $\triangleright$ Compute cosine similarity matrix $M \in \mathbb{R}^{B\times N_{b}\times K} $
  \STATE $R\leftarrow \max(M,\text{dim}=3)$ \hfill $\triangleright$ Row-wise operation $R \in \mathbb{R}^{B\times N_{b}}$ 
  \STATE $w\leftarrow \mathrm{softmax}(R)$ 
  \STATE $\text{score}\leftarrow \sum_{\text{dim}=2}(w\odot R)$ \hfill $\triangleright$ Local-level scoring $score \in \mathbb{R}^B$
  \STATE $\mathbf{M}^{l}_{\mathrm{s2t}}[b]\leftarrow \text{score}$ 
\ENDFOR
\end{algorithmic}
\end{algorithm}

\begin{figure}[htbp]
  \centering
  \includegraphics[scale=0.58]{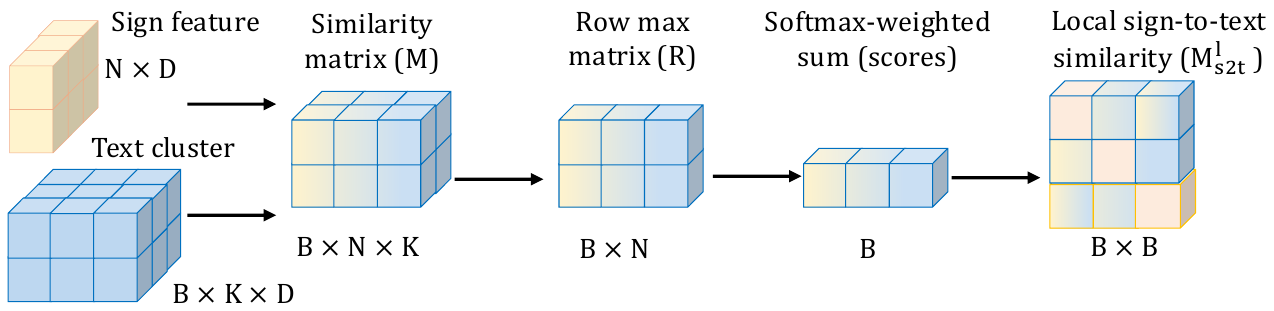}
  \caption{Illustration of the computation of our local sign-to-text cluster-wise similarity inspired by \citep{chen2020simple, radford2021learning, li2022blip, hou2024bagformer}. The similarity matrix $M$ is computed between the sign feature of each sample and all textual clusters. For each sample, the maximum similarity score is computed using a max operation, which forms $R$. The resulting values are passed through a softmax-weighted sum function to obtain the local similarity $scores$. Finally, in-batch local cluster contrastive learning is applied to pull semantically aligned visual-text pairs (highlighted in apricot) closer together, while pushing apart unaligned pairs. This process enables localised semantic grounding by focusing on the most relevant visual-text associations within each cluster.}
  \label{fig:local-cw-sim}
\end{figure}

Glosses serve as simplified representations of sign language segments in continuous video, and the extra supervision they provide has significantly improved SLU performance. However, they come with many limitations (see Appendix). Our aggregator promotes hierarchical alignment by approximating gloss-like groupings through local feature clustering. It groups subword-level textual tokens into semantically meaningful units. For instance, “curiosity” is split into “curios” and “ity,” and \begin{CJK}{UTF8}{gbsn}“背包”\end{CJK} into \begin{CJK}{UTF8}{gbsn}“背”\end{CJK} and \begin{CJK}{UTF8}{gbsn}“包”\end{CJK} by tokenizers. Both are expressed as continuous sign sequences, while each subword could be intended to align with distinct visual segments. Our aggregator dynamically merges them into more concrete phrase-level units, with the number of clusters adaptively determined by sentence structure and bounded by sentence lengths. This enables our model to preserve compositional semantics and reduce alignment errors by preventing disjoint mappings of continuous signs.

We express the computation of the local cluster-wise similarity at Algorithm \ref{alg:cluster_sim}. For brevity, Algorithm \ref{alg:cluster_sim} only presents the local sign-to-text similarity $\mathbf{M}^{l}_{\mathrm{s2t}}$, and the local text-to-sign similarity is computed analogously, with the positions of sign tokens and textual clusters exchanged. We experiment with several different row-wise operations and local-level scoring methods in the Appendix to provide a better understanding of the design choice. 

HAL is adopted to align different levels of paired features from different modalities. This dual-level strategy encourages semantically meaningful and discriminative cross-modal representations. The global and local losses are computed as follows:

\begin{equation} \label{eq:sim_loss}
\begin{split}
    & \mathcal{L}^{\phi}(S_f, T_f) = \frac{1}{2} \left( L^{\phi}_{s2t}(S_f, T_f) + L^{\phi}_{t2s}(T_f, S_f) \right) \\
    & L^{\phi}_{s2t}(S_f, T_f) = -\frac{1}{b} \sum_{i=1}^{b} \log \frac{\exp \left( \text{sim}(S^i_{f}, T^i_{f}; \phi) / \tau_{\phi} \right)}{\sum_{j=1}^{b} \exp \left( \text{sim}(S^i_{f}, T^j_{f}; \phi) / \tau_{\phi} \right)} 
\end{split}
\end{equation}

For each sign-text feature pair ($S_f^{i}$, $T_f^{i}$) in a batch $b$, we compute the bidirectional global contrastive loss $\mathcal{L}^\phi$ with temperature-scaled similarity $\tau\phi$ controlled by parameters $\phi$ (as shown Equation \ref{eq:sim_loss}). The goal is to maximise similarity for matched pairs and minimise it for mismatches ($S_f^{i}$, $T_f^{j}$), $j \neq i$. We show the sign-to-text loss $L^{\phi}_{s2t}$ explicitly; the text-to-sign loss $L^{\phi}_{t2s}$ is omitted for brevity, as it is defined in the same manner as $L^{\phi}_{s2t}$, with the roles of text and sign features reversed. The local contrastive loss follows the same structure but omits temperature scaling to emphasise sharper fine-grained alignment.

To balance both alignment levels, we introduce a parameter $\alpha \in [0, 1]$, and define the HAL loss as:

\begin{equation} \label{eq:hal}
    \mathcal{L}_{HAL} = (1-\alpha) \mathcal{L}^\phi_{global}(S_f, T_f) + \alpha \mathcal{L}^\phi_{local}(S_f, T_f)
\end{equation}



\subsubsection{Sign-grounded text matching and language modelling}

\begin{figure}[htb]
  \centering
  \includegraphics[width=0.5\textwidth]{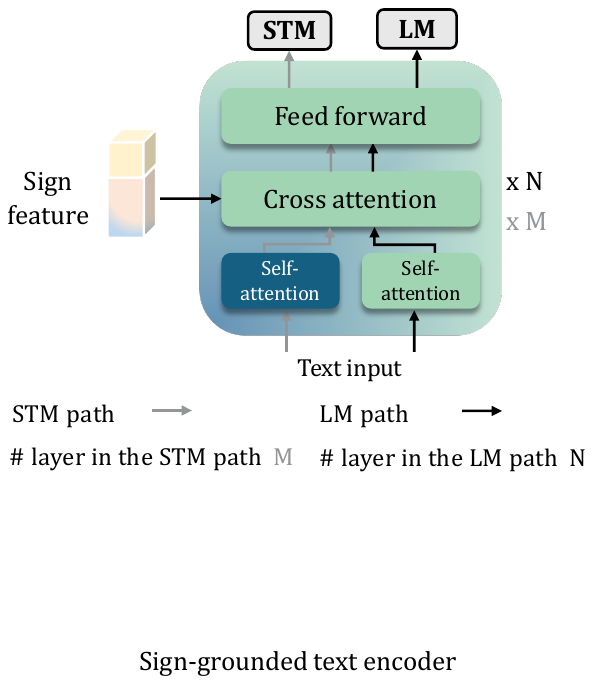}
  \caption{The architecture of the SGT encoder, which consists of two paths: the \textbf{STM} path injects sign features via cross-attention for semantic alignment, and the LM path preserves linguistic fluency through standard transformer layers (check the Appendix for details).}
  \label{fig:SGT_encoder}
\end{figure}

To improve training efficiency and foster deeper cross-modal understanding, we propose an SGT encoder, inspired by \citep{chen2020simple, radford2021learning,li2021align, li2022blip}. It unifies sign-text matching (STM) and language modelling (LM), supporting dynamic alignment of visual and linguistic features within a single framework. A task-specific token guides the model to produce multimodal embeddings. A lightweight \textit{STM} head, trained with binary cross-entropy, judges sign-text alignment. In parallel, the encoder autoregressively generates text with masked self-attention, with a cross-entropy \textit{LM} loss enhancing language structure and semantics. To balance the synergy between matching and generation, we define a composite SGT loss:
\begin{equation} \label{eq:sgt}
    \mathcal{L}_{SGT} = (1 - \beta) \mathcal{L}_{STM}(S_f, T) + \beta \mathcal{L}_{LM}(S_f, T),
\end{equation}
where $\beta \in [0,1]$ controls task emphasis. Matching enhances visual grounding, while generation regularises semantic coherence.

The pre-training objective integrates $\mathcal{L}_{HAL}$ and $\mathcal{L}_{SGT}$ as:

\begin{equation}
    \mathcal{L}_{pre\text{-}train} = \mathcal{L}_{HAL} + \mathcal{L}_{SGT}
\end{equation}


\subsection{Sign language fine-tuning}
A unified architecture is designed for all the downstream SLU tasks, casting ISLR, CSLR, and SLT as conditional language modelling. The fine-tuning objective is defined as:
\begin{equation}
    \mathcal{L}_{task} = \mathcal{L}_{LM}({T}_{out}, T_{task})
\end{equation}
where $T_{\text{out}}$ is the prediction, $T_{task}$ is the ground truth, and $task \in \{\text{ISLR, CSLR, SLT}\}$, with $T_{task}$ as a gloss (ISLR), a gloss sequence (CSLR), or a sentence (SLT).

\section{Experiment}
\textbf{Datasets}. We evaluate Sigma on a diverse set of benchmarks spanning different sign and spoken languages. WLASL2000 \citep{li2020word} is used for ISLR evaluation, CSL-Daily \citep{zhou2021improving} serves as the benchmark both for CSLR and SLT. How2Sign \citep{duarte2021how2sign} and OpenASL \citep{shi2022open} datasets are used for SLT evaluation. For large-scale pre-training, we utilise CSL-News \cite{Li2025sign} for Chinese sign language  and YouTube-ASL \cite{uthus2023youtube} for American sign language , ensuring that the pre-training data are distinct from the downstream fine-tuning benchmarks. 

\begin{table}[htbp]
\centering
\captionsetup{font=footnotesize}
\caption{Dataset statistics (Sizes in GB).}
\label{tab:datasets}
\normalsize                        
\setlength{\tabcolsep}{2pt}
\renewcommand{\arraystretch}{1.35}
\begin{tabular}{clccccc} 
\toprule
 & Dataset & Language & Level & \# Samples & Size (RGB) & Size (Skeleton) \\
\cmidrule(lr){2-7}
\multirow{2}{*}{\rotatebox{90}{Pre-training}} 
& YouTube-ASL \cite{uthus2023youtube}  & American & Sentence & 530,161 & 2,320.87 & 108.30 \\
& CSL-News \cite{Li2025sign}           & Chinese  & Sentence & 722,715 & 5,470.86 & 255.29 \\
\midrule
\multirow{4}{*}{\rotatebox{90}{Fine-tuning}} 
& WLASL \citep{li2020word}             & American & Gloss    & 21,083  & 78.84    & 3.68  \\
& How2Sign \citep{duarte2021how2sign}  & American & Sentence & 35,263  & 329.00   & 15.58 \\
& OpenASL \citep{shi2022open}          & American & Sentence & 98,419  & 638.03   & 29.78 \\
& CSL-Daily \citep{zhou2021improving}  & Chinese  & Sentence & 20,654  & 92.80    & 4.27  \\
\bottomrule
\end{tabular}
\end{table}

Table \ref{tab:datasets} summarises the benchmark sign language datasets used in this study, including language, linguistic level, number of samples, and storage size for both RGB and skeletal modalities. YouTube-ASL, WLASL, How2Sign, and OpenASL represent American sign language; CSL-News and CSL-Daily correspond to Chinese sign language. Collectively, these datasets cover both gloss-level and sentence-level annotations, enabling comprehensive evaluation across ISLR, CSLR, and SLT tasks. Although RGB videos require substantial storage capacity, skeletal representations are significantly more compact, reducing storage requirements by approximately an order of magnitude. This reduction leads to faster data loading and lower computational overhead, improving scalability during large-scale training and facilitating efficient deployment. In scenarios where users prioritise gesture dynamics over visual appearance, skeletal inputs offer a practical and resource-efficient alternative. Beyond computational efficiency, skeletal representations abstract away background clutter and appearance variations, focusing instead on body motion dynamics and structural articulation. This abstraction preserves linguistically relevant motion patterns while reducing irrelevant visual noise, which is particularly beneficial for robust SLU. As shown in Table \ref{tab:modality_comparison}, skeletal data substantially decreases both the average file size per sample and the loading time required for a single instance, further supporting its suitability for efficient model training and real-world applications.

\begin{table}[htbp]
\centering
\caption{Comparison of RGB and skeletal data.}
\label{tab:modality_comparison}
\fontsize{10pt}{10pt}\selectfont 
\begin{tabular}{ccc} 
\toprule
Modality & Avg. size per sample (KB) & Loading time per sample (ms) \\
\cmidrule(lr){1-3}
RGB       & 4714.84 & 455.35  \\
Skeletal  & 437.18  & 9.58   \\
\bottomrule
\end{tabular}
\end{table}

\textbf{Evaluation metrics}. Following prior works, we report per-class (P-C) and per-instance (P-I) Top-1 accuracy for ISLR, word error rate (WER) for CSLR, and BLEU \& ROUGE-L scores for SLT. For brevity, we denote BLEU-1, BLEU-4, and ROUGE-L as B@1, B@4, and R@L in the tables of the following sections. 

\begin{table}[htp]
\centering
\caption{Training settings across tasks.}
\fontsize{10pt}{10pt}\selectfont 
\begin{tabular}{clccc} 
\toprule
Phase & Settings & ISLR & CSLR & SLT \\
\midrule

\multirow{4}{*}{Shared}
& Optimiser & \multicolumn{3}{c}{AdamW} \\
& Weight decay & \multicolumn{3}{c}{1.00E-03} \\
& optimiser momentum & \multicolumn{3}{c}{$\beta_1,\beta_2 = 0.9, 0.999$} \\
& Learning rate schedule & \multicolumn{3}{c}{Cosine decay} \\
\midrule

\multirow{3}{*}{Pre-training}
& Training epochs & 15 & 20 & 25 \\
& Batch size & \multicolumn{3}{c}{16} \\
& Learning rate & \multicolumn{3}{c}{1.00E-04} \\
\midrule

\multirow{3}{*}{Fine-tuning}
& Training epochs & 10 & 15 & 15 \\
& Batch size & \multicolumn{3}{c}{8} \\
& Learning rate & 1.00E-07 & \multicolumn{2}{c}{1.00E-06} \\
\bottomrule
\end{tabular}
\label{tab:training_settings}
\end{table}

\textbf{Training details}. The training settings are configured and listed in Table \ref{tab:training_settings}.

\section{Comparison with state-of-the-art methods}
We evaluate Sigma across the five aforementioned core SLU tasks. For ISLR on the WLASL2000 dataset, our model sets a new performance benchmark (see Table \ref{tab:ISRL_CSLR_results} (a)). These results demonstrate strong gesture recognition and effective feature discrimination. For CSLR on the CSL-Daily dataset, as shown in Table \ref{tab:ISRL_CSLR_results} (b), Our method achieves new state-of-the-art (SOTA) performance, surpassing the strong pose-RGB-based Uni-Sign model, highlighting improved temporal modelling and more precise alignment between sign sequences and sign glosses relying solely on skeletal data. For SLT (see Table \ref{tab:openasl_result}, Table \ref{tab:how2sign_result}, and Table \ref{tab:csl_result}), Sigma shows strong performance across How2Sign, OpenASL, and CSL-Daily. On How2Sign, it delivers improvements across all evaluation metrics. Sigma achieves new SOTA results on OpenASL across all evaluation metrics used for this study. On CSL-Daily, it surpasses all gloss-free methods and rivals the long-standing gloss-based SOTA model CV-SLT. These results confirm the generalisability of Sigma across varied datasets and SLU tasks. 


\begin{table}[htbp]
\centering
\caption{Performance comparison of ISLR and CSLR \textbf{(a)} ISLR results on the WLASL2000 dataset evaluated by Per-Instance (P-I) and Per-Class (P-C) accuracy. \textbf{(b)} CSLR results on the CSL-Daily dataset evaluated by Word Error Rate (WER).}
\label{tab:ISRL_CSLR_results}

\begin{minipage}[t]{0.48\textwidth}
\centering
{\footnotesize (a) ISLR results on WLASL2000} \\[1ex]
{\footnotesize
\fontsize{10pt}{10pt}\selectfont 
\begin{tabular}{lcc}
\toprule
\multirow{2}{*}{Method} & \multicolumn{2}{c}{TEST} \\ 
\cmidrule(lr){2-3}
& P-I$\uparrow$ & P-C$\uparrow$ \\
\midrule
ST-GCN \citep{yan2018spatial} & 34.40 & 32.53 \\
HMA \citep{hu2021hand} & 37.91 & 35.90 \\
SignBERT \citep{zhou2021signbert} & 39.40 & 36.74 \\
BEST \citep{zhao2023best} & 46.25 & 43.52 \\
SignBERT+ \citep{hu2023signbert+} & 48.85 & 46.37 \\
MSLU \citep{zhou2024scaling} & 56.29 & 53.29 \\
NLA-SLR \citep{zuo2023natural} & 61.05 & 58.05 \\
Uni-Sign \citep{Li2025sign} & 63.52 & 61.32 \\
\midrule
\textbf{Sigma} & \textbf{64.54} & \textbf{62.28} \\
\bottomrule
\end{tabular}
}
\end{minipage}\hfill
\begin{minipage}[t]{0.48\textwidth}
\centering
{\footnotesize (b) CSLR results on CSL-Daily} \\[1ex]
\renewcommand{\arraystretch}{1.13}
{\footnotesize
\fontsize{10pt}{10pt}\selectfont 
\begin{tabular}{lcc} 
\toprule
\multirow{2}{*}{Method} & \multicolumn{1}{c}{DEV} & \multicolumn{1}{c}{TEST} \\
\cmidrule(lr){2-2} \cmidrule(lr){3-3}
                        & WER$\downarrow$ & WER$\downarrow$ \\
\midrule
SignBT \citep{zhou2021improving} & 33.20 & 33.20 \\
AdaBrowse \citep{hu2023adabrowse} & 31.20 & 30.70 \\
SEN \citep{hu2023self} & 31.10 & 30.70 \\
CorrNet \citep{hu2023continuous} & 30.60 & 30.10 \\
MSLU \citep{zhou2024scaling} & 28.60 & 27.90 \\
CoSign \citep{jiao2023cosign} & 28.10 & 27.20 \\
Uni-Sign \citep{Li2025sign} & 26.70 & 26.00 \\
\midrule
\textbf{Sigma} & \textbf{26.09} & \textbf{25.26} \\
\bottomrule
\end{tabular}
}
\end{minipage}

\end{table}

\begin{table}[htbp]
\centering
\captionsetup{font=footnotesize}
\caption{SLT results on OpenASL dataset (Full results, including B@2 and B@3, are in the Appendix).}
\fontsize{10pt}{12pt}\selectfont
\begin{tabular}{lcccccc} 
\toprule
\multirow{2}{*}{Method} & \multicolumn{3}{c}{DEV} & \multicolumn{3}{c}{TEST} \\ 
\cmidrule(r){2-4}\cmidrule(l){5-7}
& B@1$\uparrow$ & B@4$\uparrow$ & R@L$\uparrow$
& B@1$\uparrow$ & B@4$\uparrow$ & R@L$\uparrow$ \\ 
\midrule
GloFE-VN \citep{lin2023gloss}
& 21.06 & 6.68  & 21.37
& 21.56 & 7.06  & 21.75 \\
Conv-GRU \citep{camgoz2018neural}
& 16.72 & 4.82  & 16.25
& 16.11 & 4.58  & 16.10 \\
I3D-transformer \citep{shi2022open}
& 18.26 & 5.60  & 18.88
& 18.31 & 5.56  & 18.64 \\
OpenASL \citep{shi2022open}
& 20.10 & 6.57  & 20.43
& 20.92 & 6.72  & 21.02 \\
Uni-Sign \citep{Li2025sign}
& 50.84 & 24.16 & 44.58
& 49.35 & 23.14 & 43.22 \\
$C^{2}$RL \citep{chen2025c}
& -     & -     & -
& 31.46 & 13.21 & 31.36 \\
\textbf{Sigma}
& \textbf{51.68} & \textbf{25.72} & \textbf{45.81}
& \textbf{49.91} & \textbf{23.21} & \textbf{45.38} \\
\bottomrule
\end{tabular}
\label{tab:openasl_result}
\end{table}

\begin{table}[htb]
\centering
\caption{SLT results on How2Sign dataset (Full results, including B@2 and B@3, are in the Appendix).}
\fontsize{10pt}{10pt}\selectfont 
\begin{tabular}{lccccc} 
\toprule
\multirow{2}{*}{Method} & \multicolumn{5}{c}{TEST}                                                                                                                   \\ 
\cmidrule(r){2-6}
                        & B@1$\uparrow$ & B@2$\uparrow$ & B@3$\uparrow$ & B@4$\uparrow$ & R@L$\uparrow$ \\ 
\toprule

GloFE-VN \citep{lin2023gloss}
& 14.90 & 7.30 & 3.90 & 2.20 & 12.60 \\

YouTube-ASL \citep{uthus2023youtube}
& 37.80
& 24.10
& 16.90
& 12.40
& - \\

MSLU \citep{zhou2024scaling}
& 20.10 & 7.70 & - & 2.40 & 17.20 \\


$C^{2}RL$ \citep{chen2025c}
& 29.10 & 18.60 & 12.90 & 9.40 & 27.00 \\

FLa-LLM \citep{chen2024factorized}
& 29.80 & 19.00 & 13.30 & 9.70 & 27.80 \\

\textbf{Sigma}
& \textbf{39.89}
& \textbf{27.12}
& \textbf{19.81}
& \textbf{14.92}
& \textbf{36.88} \\

\bottomrule
\end{tabular}
\label{tab:how2sign_result}
\end{table}

\begin{table}[h!]
\centering
\caption{SLT results on CSL-Daily dataset (Full results, including B@2 and B@3, are in the Appendix).}
\fontsize{10pt}{12pt}\selectfont 
\begin{tabular}{clcccccc} 
\toprule
 & Method 
& \multicolumn{3}{c}{DEV}
& \multicolumn{3}{c}{TEST} \\
\cmidrule(lr){3-5}\cmidrule(lr){6-8}
& 
& B@1$\uparrow$ & B@4$\uparrow$ & R@L$\uparrow$
& B@1$\uparrow$ & B@4$\uparrow$ & R@L$\uparrow$ \\
\midrule

\multirow{7}{*}{\rotatebox{90}{Gloss-based}}
& SLRT \citep{camgoz2020sign}
& 37.47 & 11.88 & 37.96
& 37.38 & 11.79 & 36.74 \\


& SignBT \citep{zhou2021improving}
& 51.46 & 20.80 & 49.49
& 51.42 & 21.34 & 49.31 \\

& MMTLB \citep{chen2022simple}
& 53.81 & 24.42 & 53.38
& 53.31 & 23.92 & 53.25 \\

& SLTUNET \citep{zhang2023sltunet}
& - & 23.99 & 53.58
& 54.98 & 25.01 & 54.08 \\

& TS-SLT \citep{chen2022two}
& 55.21 & 25.76 & 55.10
& 55.44 & 25.79 & 55.72 \\

& CV-SLT \citep{zhao2024conditional}
& 58.05 & 28.24 & 56.36
& 58.29 & 28.94 & 57.06 \\

\midrule

\multirow{11}{*}{\rotatebox{90}{Gloss-free}}
& SLRT \citep{camgoz2020sign}
& 21.03 & 4.04 & 20.51
& 20.00 & 3.03 & 19.67 \\

& GASLT \citep{yin2023gloss}
& - & - & -
& 19.90 & 4.07 & 20.35 \\

& MSLU \citep{zhou2024scaling}
& 33.28 & 10.27 & 33.13
& 33.97 & 11.42 & 33.80 \\

& NSLT \citep{camgoz2018neural}
& 34.22 & 7.96 & 34.28
& 34.16 & 7.56 & 34.54 \\

& GFSLT-VLP \citep{zhou2023gloss}
& 39.20 & 11.07 & 36.70
& 39.37 & 11.00 & 36.44 \\

& FLa-LLM \citep{chen2024factorized}
& - & - & -
& 37.13 & 14.20 & 37.25 \\

& $C^{2}RL$ \citep{chen2025c}
& - & - & -
& 49.32 & 21.61 & 48.21 \\

& Uni-Sign \citep{Li2025sign}
& 55.30 & 26.25 & 56.03
& 55.08 & 26.36 & 56.51 \\

& SignLLM \citep{gong2024llms}
& 42.45 & 12.23 & 39.18
& 39.55 & 15.75 & 39.91 \\

& Sign2GPT \citep{wong2024sign2gpt}
& - & - & -
& 41.75 & 15.40 & 42.36 \\

& \textbf{Sigma}
& \textbf{56.12} & \textbf{27.80} & \textbf{57.12}
& \textbf{55.73} & \textbf{27.12} & \textbf{57.32} \\

\bottomrule
\end{tabular}
\label{tab:csl_result}
\end{table}

\section{Qualitative analysis}
To further validate the semantic advantages of our proposed method, we present qualitative results derived for all benchmark datasets used in this study. Each table contrasts ground-truth references with results outputted by our method (Sigma). 

\begin{figure}[htbp]
  \centering
  \includegraphics[scale=0.88]{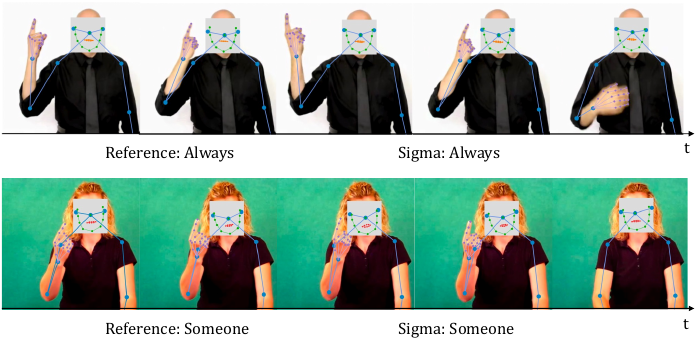}
  \caption{Qualitative examples derived from the WLASL2000 dataset for ISLR.}
  \label{fig:quli_islr}
\end{figure}

For ISLR, Figure \ref{fig:quli_islr} presents qualitative examples from the WLASL2000 dataset. Sigma demonstrates consistent recognition across signers with varying visual appearances and signing styles. This stability reflects the improved semantic grounding and more effective cross-modal alignment, which together help mitigate influence caused by subtle differences in gesture execution. By capturing both fine-grained motion details and broader temporal structure, Sigma supports more reliable recognition, aligning with our objective of balancing local precision with global contextual understanding.

In many examples within each table, our method achieves complete correctness, perfectly replicating references. Although these illustrate baseline competence, the deeper value of our framework emerges in the challenging examples—those where slight variations occur between the predictions of the model and the ground-truth. Instead of treating these discrepancies as outright errors, we analyze them through the lens of semantic grounding and linguistic alignment.

For CSLR, as shown in the bottom rows of Table \ref{tab:qualitative_all} (a), Sigma exhibits semantic preservation despite minor lexical variations. For example, the predicted phrase \begin{CJK}{UTF8}{gbsn}“固定 封面”\end{CJK} (fix, cover) differs in wording from the reference \begin{CJK}{UTF8}{gbsn}“决定 表面”\end{CJK} (set, surface), but within the specific visual context, they convey similar semantic intent. While such variations reduce exact-match scores like the WER, they are easily understood by human readers, as language naturally allows multiple ways to express similar visual concepts. These cases highlight the enhanced semantic grounding: even when the predicted glosses deviate from the reference, the core meaning remains largely intact. This is especially crucial in CSLR, where explicit gloss segmentation is absent and contextual understanding plays a key role.

\begin{CJK}{UTF8}{gbsn}
\begin{center}
\captionsetup{font=footnotesize}
\captionof{table}{Qualitative examples for CSLR and SLT benchmarks. For Chinese datasets (CSL-Daily), English translations are provided in parentheses below the original text.}
\label{tab:qualitative_all}
\begin{longtable}{c p{0.38\textwidth} p{0.38\textwidth}}
\toprule
& Reference & Sigma (Predicted) \\
\midrule
\endfirsthead
\toprule
\endhead
\bottomrule
\endfoot

\multirow{12}{*}{\rotatebox{90}{\begin{tabular}{c}(a)\\ CSL-Daily (CSLR)\end{tabular}}}
& 椅子 他们 想 什么 时间 去 买
& 椅子 他们 想 什么 时间 去 买 \\
& \textit{(Chairs, they, want, what, time, to go, buy)}
& \textit{(Chairs, they, want, what, time, to go, buy)} \\
& 帮助 看 这 衣服 怎么样
& 帮助 看 这 衣服 怎么样 \\
& \textit{(Help, look, these, clothes, how)}
& \textit{(Help, look, these, clothes, how)} \\
& 可以 这 近 不 远 饭店 走 多少 到
& 可以 这 近 不 远 饭店 走 多少 到 \\
& \textit{(Okay, here, near, not, far, restaurant, walk, how long, arrive)}
& \textit{(Okay, here, near, not, far, restaurant, walk, how long, arrive)} \\
& 计算 结果 我们 必须 要 准确
& 计算 结果 我们 必须 准确 \\
& \textit{(Calculate, result, we, must, need, accurate)}
& \textit{(Calculate, result, we, must, accurate)} \\
& 存折 丢 快 去 银行 增加 办
& 你 存折 丢 快 去 银行 增加 办理 \\
& \textit{(Bankbook, lose, quickly, go, bank, add, apply)}
& \textit{(You, bankbook, lose, quickly, go, bank, add, apply)} \\
& 核磁共振 方法 来 决定 表面 机器
& 核磁共振 方法 来 固定 封面 机器 \\
& \textit{(MRI, method, to, set, surface, machine)}
& \textit{(MRI, method, to, fix, cover, machine)} \\
\midrule

\multirow{12}{*}{\rotatebox{90}{\begin{tabular}{c}(b)\\ CSL-Daily (SLT)\end{tabular}}}
& 我 每 天 六 点 起 床 。
& 我 每 天 六 点 起 床 。\\
& \textit{(I get up at six o'clock every day.)}
& \textit{(I get up at six o'clock every day.)} \\
& 警 察 要 检 查 你 的 身 份 证 。
& 警 察 要 检 查 你 的 身 份 证 。\\
& \textit{(The police need to check your ID card.)}
& \textit{(The police need to check your ID card.)} \\
& 苹 果 是 你 买 的 吗 ?
& 苹 果 是 你 买 的 吗 ?\\
& \textit{(Did you buy the apples?)}
& \textit{(Did you buy the apples?)} \\
& 他 醒 来 时 发 现 自 己 在 医 院 里 。
& 他 醒 来 后 发 现 自 己 在 医 院 。\\
& \textit{(When he woke up, he found himself in the hospital.)}
& \textit{(After waking up, he found himself in the hospital.)} \\
& 我 的 护 照 忘 记 带 了 , 需 要 回 家 拿 。
& 我 的 护 照 忘 带 了 , 回 家 要 拿 。\\
& \textit{(I forgot to bring my passport, I need to go home to get it.)}
& \textit{(I forgot my passport, I have to get it at home.)} \\
& 桌 上 放 着 很 多 饮 料 , 你 喝 什 么 ?
& 桌 子 上 有 很 多 饮 料 , 你 想 喝 什 么 ?\\
& \textit{(There are many drinks on the table, what will you drink?)}
& \textit{(There are many drinks on the table, what do you want to drink?)} \\
\midrule

\multirow{6}{*}{\rotatebox{90}{\begin{tabular}{c}(c)\\ How2Sign (SLT)\end{tabular}}}
& My name is Dr. Art Bowler.              & My name is Dr. Art Bowler. \\
& What do you see?                        & What do you see? \\
& My name is Allen Diwan.                 & Hi, I'm Allen Diwan. \\
& You're having a good time along the way. & It's a really enjoyable process. \\
& Stay safe, and we'll see ya' next time.  & See you next time. \\
& I hope you're having fun.               & I hope you had fun with it. \\
\midrule

\multirow{6}{*}{\rotatebox{90}{\begin{tabular}{c}(d)\\ OpenASL (SLT)\end{tabular}}}
& America!                                        & America! \\
& I'm from Austin, Texas!                         & I'm from Austin, Texas! \\
& See you at the conference this July!            & See you at the conference this July! \\
& That's not right!                               & That's not fair. \\
& Also, be sure you talk to your legislators.     & You will also talk with your legislators. \\
& Progress is being made.                         & The work is moving forward. \\

\end{longtable}
\end{center}
\end{CJK}

For SLT, qualitative examples from English-based datasets (Tables \ref{tab:qualitative_all} (b)-(c)) demonstrate the ability of the model to produce fluent and contextually appropriate sentences, as well as the generalization of the model across speaker identities and conversational styles. For instance, in Table \ref{tab:qualitative_all} (b), the model converts “I hope you’re having fun.” into “I hope you had fun with it.” showing an understanding of tense and implied context. These changes demonstrate that the model captures deeper semantic meaning rather than relying only on surface-level similarity. In Table \ref{tab:qualitative_all} (c), the model outputs “You will also talk with your legislators” instead of the reference “Also, be sure you talk to your legislators.” Though not a verbatim match, the generated sentence is syntactically sound and preserves the core message, demonstrating sentence-level comprehension. This reflects the impact of our SignEF strategy in bridging local-global semantics across modalities. 

In Table \ref{tab:qualitative_all} (d), the Chinese SLT examples show similar robust results. In the fourth row, the reference \begin{CJK}{UTF8}{gbsn}“他醒来时发现自己在医院里”\end{CJK} (When he woke up, he found himself in the hospital) and the prediction \begin{CJK}{UTF8}{gbsn}“他醒来后发现自己在医院”\end{CJK} (After waking up, he found himself in the hospital) use slightly different syntax but express the exact same idea. In the fifth example, the phrases \begin{CJK}{UTF8}{gbsn}“需要回家拿”\end{CJK} (need to go home to get it) and \begin{CJK}{UTF8}{gbsn}“回家要拿”\end{CJK} (have to get it at home) describe the same action with a minor variation in sentence structure. In the final row, the model substitutes \begin{CJK}{UTF8}{gbsn}“你喝什么?”\end{CJK} (what will you drink?) with \begin{CJK}{UTF8}{gbsn}“你想喝什么?”\end{CJK} (what do you want to drink?), a natural phrasing that enhances the sentence without altering the core meaning. While these lexical variations negatively impact strict token-matching metrics like BLEU or ROUGE, they accurately reflect natural language usage and maintain high communicative clarity.

\begin{center}
\captionsetup{font=footnotesize}
\captionof{table}{Comprehensive ablation study of Sigma. We analyse the impact of (a) pre-training data scale, (b) SignEF fusion depth, (c) the local–global balancing weight $\alpha$ in $\mathcal{L}_{HAL}$, (d) the trade-off weight $\beta$ in $\mathcal{L}_{SGT}$, and (e) different pre-training strategies. Results are reported on WLASL2000 (ISLR) and CSL-Daily (CSLR and SLT). \colorbox{darkb}{Best} results and \colorbox{lightb}{second best} results are highlighted.}
\label{tab:main_ablation}
\renewcommand{\arraystretch}{0.8}
\begin{longtable}{clcccccc}
\toprule
& & \multicolumn{2}{c}{WLASL2000} 
  & \multicolumn{4}{c}{CSL-Daily} \\
\cmidrule(lr){3-4} \cmidrule(lr){5-8}
& Setting 
& \multicolumn{2}{c}{ISLR} 
& CSLR 
& \multicolumn{3}{c}{SLT} \\
\cmidrule(lr){3-4} \cmidrule(lr){5-5} \cmidrule(lr){6-8}
& 
& P-I $\uparrow$ & P-C $\uparrow$ 
& WER $\downarrow$ 
& B@1 $\uparrow$ & B@4 $\uparrow$ & R@L $\uparrow$ \\
\midrule
\endfirsthead
\toprule        
\endhead
\endfoot
\bottomrule
\endlastfoot

\multirow{5}{*}{\rotatebox{90}{\begin{tabular}{c}(a)\\ Pre-train Scale\end{tabular}}}
& 0\%   & 23.00 & 22.21 & 72.34 & 19.87 & 4.12  & 20.18 \\
& 25\%  & 50.68 & 48.92 & 32.18 & 43.78 & 19.23 & 47.58 \\
& 50\%  & 58.83 & 56.80 & 29.78 & 50.81 & 23.99 & 53.12 \\
& 75\%  & \cellcolor{lightb}62.46 & \cellcolor{lightb}60.29 
        & \cellcolor{lightb}27.92 & \cellcolor{lightb}53.92 
        & \cellcolor{lightb}26.12 & \cellcolor{lightb}56.12 \\
& 100\% & \cellcolor{darkb}\textbf{64.54}
         & \cellcolor{darkb}\textbf{62.28}
         & \cellcolor{darkb}\textbf{25.26}
         & \cellcolor{darkb}\textbf{55.73}
         & \cellcolor{darkb}\textbf{27.12}
         & \cellcolor{darkb}\textbf{57.32} \\
\midrule

\multirow{6}{*}{\rotatebox{90}{\begin{tabular}{c}(b)\\ Fusion depth\end{tabular}}}
& 0 & 62.12 & 60.23 & 26.89 & 54.21 & 25.12 & 55.17 \\
& 1 & 62.90  & 60.50  & 27.10  & 54.80 & 25.90  & 55.80 \\
& 2 & 63.60  & 61.30  & 26.00  & 55.40 & 26.60  & 56.80 \\
& 3 & 64.10  & 61.90  
      & \cellcolor{darkb}\textbf{25.26}
      & \cellcolor{darkb}\textbf{55.73}
      & \cellcolor{darkb}\textbf{27.12}
      & \cellcolor{darkb}\textbf{57.32} \\
& 4 & \cellcolor{lightb}64.35 
      & \cellcolor{lightb}62.10 
      & \cellcolor{lightb}25.40
      & \cellcolor{lightb}55.60
      & \cellcolor{lightb}27.10
      & \cellcolor{lightb}57.10 \\
& 5 & \cellcolor{darkb}\textbf{64.54}
      & \cellcolor{darkb}\textbf{62.28}
      & 25.55 & 55.0 & 27.01 & 57.13 \\
\midrule

\multirow{7}{*}{\rotatebox{90}{\begin{tabular}{c}(c)\\ $\alpha$ in $\mathcal{L}_{HAL}$\end{tabular}}}
& 0.0 & 63.12 & 61.57 & 26.68 & 54.79 & 26.32 & 56.21 \\
& 0.2 & 64.20 & 62.00 & 26.10 & 55.20 & 26.80 & 56.90 \\
& 0.4 & \cellcolor{lightb}64.40 
      & \cellcolor{lightb}62.20 
      & \cellcolor{lightb}25.70 
      & \cellcolor{lightb}55.55 
      & 27.00 
      & \cellcolor{lightb}57.15 \\
& 0.5 & \cellcolor{darkb}\textbf{64.54}
      & \cellcolor{darkb}\textbf{62.28}
      & \cellcolor{darkb}\textbf{25.26}
      & \cellcolor{darkb}\textbf{55.73}
      & \cellcolor{lightb}\textbf{27.12}
      & \cellcolor{darkb}\textbf{57.32} \\
& 0.6 & 64.38 & 62.18 & 25.80 & 55.60 & \cellcolor{darkb}\textbf{27.16} & 57.20 \\
& 0.8 & 64.10 & 61.95 & 26.30 & 54.89 & 26.70 & 56.80 \\
& 1.0 & 63.91 & 61.71 & 26.78 & 54.12 & 26.21 & 56.13 \\
\midrule

\multirow{7}{*}{\rotatebox{90}{\begin{tabular}{c}(d)\\ $\beta$ in $\mathcal{L}_{SGT}$ \end{tabular}}}
& 0.0 & 63.80 & 61.73 & 26.43 & 54.80 & 26.12 & 55.85 \\
& 0.2 & 64.10 & 61.90 & 26.40 & 55.11 & 26.70 & 56.91 \\
& 0.4 & \cellcolor{lightb}64.35 
      & \cellcolor{lightb}62.05 
      & \cellcolor{lightb}25.90 
      & \cellcolor{lightb}55.51 
      & \cellcolor{darkb}\textbf{27.27} 
      & \cellcolor{lightb}57.15 \\
& 0.5 & 64.41 & 62.15 
      & \cellcolor{darkb}\textbf{25.26}
      & \cellcolor{darkb}\textbf{55.73}
      & \cellcolor{lightb}27.12 
      & \cellcolor{darkb}\textbf{57.32} \\
& 0.6 & \cellcolor{darkb}\textbf{64.54}
      & \cellcolor{darkb}\textbf{62.28}
      & 25.60 & 55.60 & 27.05 & 57.28 \\
& 0.8 & 64.20 & 62.00 & 26.10 & 55.20 & 26.80 & 56.95 \\
& 1.0 & 63.12 & 61.43 & 26.79 & 54.56 & 26.43 & 56.12 \\
\midrule

\multirow{4}{*}{\rotatebox{90}{\begin{tabular}{c}(e)\\ Training \end{tabular}}}
& None          & 0.02  & 0.01  & 523.85 & 7.14  & 0.12  & 11.95 \\
& Decoder only  & 1.80  & 1.50  & 150    & 18    & 3.10  & 18.50 \\
& Encoder only  & \cellcolor{lightb}28.40 
                & \cellcolor{lightb}26.32 
                & \cellcolor{lightb}27.91 
                & \cellcolor{lightb}53.81 
                & \cellcolor{lightb}25.62 
                & \cellcolor{lightb}54.61 \\
& Full          & \cellcolor{darkb}\textbf{64.54}
                & \cellcolor{darkb}\textbf{62.28}
                & \cellcolor{darkb}\textbf{25.26}
                & \cellcolor{darkb}\textbf{55.73}
                & \cellcolor{darkb}\textbf{27.12}
                & \cellcolor{darkb}\textbf{57.32} \\

\end{longtable}
\end{center}

\section{Ablation study}
In this section, we conduct several ablation studies to verify the effectiveness of our proposed framework. 

\subsection{Impact of pre-training data scale}
To examine the role of pre-training scale, we gradually increase the proportion of pre-training data from 0\% to 100\% (Table \ref{tab:main_ablation} (a)). Without pre-training (0\%), the model fails to establish meaningful cross-modal representations, resulting in severe degradation across all tasks, particularly CSLR (WER 72.34) and SLT (B@4 4.12). Introducing only 25\% of the data dramatically improves performance, reducing CSLR WER to 32.18 and increasing SLT B@4 to 19.23. This large jump indicates that even limited cross-modal supervision significantly enhances semantic grounding. As the scale increases further (50\% → 75\% → 100\%), performance improves consistently across ISLR, CSLR, and SLT. The gains gradually saturate but remain stable, with the best results achieved at full scale (100\%): P-I 64.54, WER 25.26, and B@4 27.12. Importantly, improvements are monotonic across all evaluation metrics, suggesting that the benefit does not stem from overfitting or optimisation artefacts, but from progressively stronger visual–linguistic alignment. These results confirm that large-scale cross-modal pre-training is the primary driver of transferable SLU representations.

\subsection{Impact of sign-aware early fusion}
We analyse the effect of SignEF by varying the fusion depth (Table \ref{tab:main_ablation} (b)), i.e., the number of layers incorporating cross-modal interaction. When fusion depth is set to 0 (no SignEF), performance is clearly lower (WER 26.89, B@4 25.12), indicating that purely late interaction is insufficient for strong alignment. Introducing shallow fusion (depth 1-2) gradually improves performance, reducing WER from 27.10 to 26.00 and increasing translation quality. Performance peaks at depth 3, achieving the lowest CSLR WER (25.26) and strongest SLT performance (B@4 27.12, R@L 57.32). Increasing fusion depth further (4-5) continues to slightly improve ISLR accuracy, but CSLR and SLT performance no longer improve and even slightly decline. This pattern suggests that early cross-modal interaction is crucial for injecting semantic priors into visual representations, but overly deep fusion may blur modality-specific characteristics or introduce redundant coupling. A moderate fusion depth therefore provides the best balance between semantic enrichment and structural stability.

\subsection{Impact of local-global feature balancing} 
We vary $\alpha$ in $\mathcal{L}_{HAL}$ to control the balance between global alignment and local cluster-wise alignment (Table~\ref{tab:main_ablation} (c)). Recall that the local branch does not operate on raw subword tokens, but on clustered semantic units that approximate gloss-like segments. These clusters merge fragmented tokenizer outputs into coherent phrase-level representations, providing more structured alignment targets for visual tokens. When $\alpha=0.0$ (pure global alignment), performance drops noticeably (WER 26.68, B@4 26.32). In this setting, alignment is enforced only at the sentence level, without explicit supervision on how visual segments correspond to intermediate semantic units. As a result, the model may capture coarse semantic similarity while missing compositional structure, leading to weaker translation quality. Conversely, when $\alpha=1.0$ (pure local alignment), WER increases to 26.78 and SLT metrics decline. Although cluster-level correspondence is strengthened, the absence of global sentence-level alignment disrupts overall semantic coherence. This indicates that clustered units alone cannot fully constrain sentence meaning without holistic guidance. The best overall performance is achieved at $\alpha=0.5$, where global and cluster-based local alignment are equally weighted (WER 25.26, B@4 27.12). In this regime, clustered textual units act as gloss-like anchors that guide fine-grained visual alignment, while global contrast preserves sentence-level semantics. Notably, slightly stronger local emphasis ($\alpha=0.6$) yields the highest B@4 (27.16), suggesting that translation particularly benefits from more precise segment-level grounding. However, ISLR and CSLR slightly decline, reflecting the need for stable global structure in recognition tasks. These results demonstrate that clustering is not merely a textual reorganisation, but an effective mechanism for restructuring cross-modal supervision. By converting fragmented subword tokens into semantically coherent alignment targets, the model reduces disjoint mappings between continuous sign motions and arbitrary tokenizer boundaries. The clear sensitivity of performance to $\alpha$ confirms that cluster-level alignment materially improves visual-text correspondence. Effective SLU therefore requires hierarchical modelling that integrates gloss-like semantic grouping with sentence-level alignment.

\subsection{Trade-off analysis between text-matching and language modelling} 
We vary $\beta$ in $\mathcal{L}_{SGT}$ to examine the trade-off between sign-text matching (STM) and language modelling (LM) (Table \ref{tab:main_ablation} (d)). At $\beta=0.0$ (pure matching), performance is clearly lower (WER 26.43, B@4 26.12), suggesting that alignment alone cannot ensure coherent linguistic generation.
At $\beta=1.0$ (pure language modelling), both ISLR and CSLR degrade (P-I 63.12, WER 26.79), indicating weakened cross-modal grounding. The optimal balance occurs around $\beta=0.5$, achieving the lowest WER (25.26) and strongest overall SLT performance (B@4 27.12, R@L 57.32). Notably, $\beta=0.4$ yields the highest B@4 (27.47), but slightly weaker recognition accuracy, reflecting a trade-off between generation fluency and discriminative grounding. These results confirm that STM and LM play complementary roles: STM strengthens cross-modal semantic alignment, while LM regularises linguistic structure. A balanced objective ensures both accurate recognition and fluent translation.

\subsection{Contribution of the core components}
Finally, we analyse the contribution of each training component. When bypassing both pre-training and fine-tuning to perform direct inference, performance collapses across all tasks, highlighting the necessity of semantically informative representation learning. Training only the decoder yields limited improvement, suggesting that language modelling alone cannot compensate for weak visual grounding. Training only the encoder significantly improves CSLR and SLT, confirming the importance of semantically enriched visual representations. Training the full framework achieves the best results across all metrics, demonstrating the synergistic effect of the following:

\begin{enumerate}
    \item Sign-aware early fusion for cross-modal enrichment,
    \item Hierarchical alignment for balanced local-global modelling,
    \item Unified STM + LM training for semantic coherence.
\end{enumerate}

These results validate the necessity of jointly optimising visual grounding, hierarchical alignment, and generative regularisation within a unified skeleton-based pre-training framework.


\section{Conclusion}
In this work, we presented Sigma, a unified skeleton-based framework that addresses weak semantic grounding, local-global feature imbalance, and inefficient cross-modal learning in SLU. By relying solely on skeletal data, Sigma abstracts away visual noise, proving that kinematic representations, when properly aligned with linguistic semantics, are highly effective for SLU. Sigma introduces: 1) the SignEF mechanism for bidirectional visual-textual interaction, 2) the HAL strategy for optimising local and global alignment via contrastive objectives, and 3) a unified pre-training scheme combining contrastive learning, text matching, and language modelling. Sigma achieves SOTA or competitive performance across ISLR, CSLR, and SLT tasks, effectively bridging the gap between isolated gesture recognition and continuous, gloss-free translation.

Despite its strong performance, Sigma has limitations. Gloss annotations improve recognition performance but constrain scalability, as annotating them is time-consuming and requires domain expertise. While recent methods attempt to reduce gloss supervision, no optimal solution currently exists that fully replaces annotations without compromising performance across all SLU tasks. Additionally, while our framework unifies multiple tasks, task-specific methods may still outperform it in certain scenarios.

Looking forward, we plan to incorporate additional modalities such as RGB and depth to provide richer visual information, while exploring balanced solutions that maintain computational efficiency. Transitioning from 2D keypoints to 3D parametric body models such as SMPL-X could provide richer spatial context and better robustness against self-occlusion. Finally, given Sigma's foundation on the multilingual mT5 architecture, extending the framework to zero-shot or few-shot multilingual sign language translation presents an exciting opportunity for advancing universal, barrier-free communication.

\section{Generative AI declaration}
During the preparation of this work, the authors used generative AI tools to assist with language editing and grammar refinement. The authors reviewed and edited the content as needed and take full responsibility for the content of the publication.

\section{Border impacts}
By advancing the performance and generalizability of SLU, our research supports more seamless interaction between the impaired community and the general public. Ultimately, this contributes to mitigating communication barriers, fostering greater inclusivity as well as equal access across social, educational, and professional environments.

\section{Motivation} \label{sec:motivation}
\begin{figure}[htbp]
  \centering
  \includegraphics[scale=1]{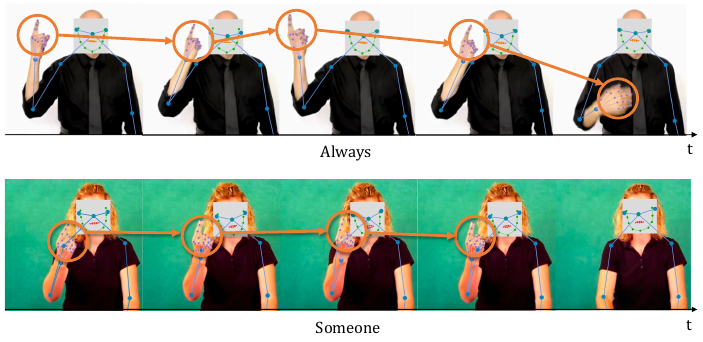}
  \caption{Visualization derived from the WLASL2000 dataset. The right hand, along with its primary motion trajectory, is highlighted to illustrate the gesture dynamics. The figure shows two sign sequences, “Always” and “Someone.” Although both gestures exhibit similar hand shapes and motion trajectories, they differ in spatial and temporal extent. Disambiguating them requires not only local visual detail but also global temporal understanding and accurate alignment with linguistic meaning, highlighting the need for effective  multimodal representation learning.}
  \label{fig:vis_wlasl}
\end{figure}

\begin{figure}[htbp]
  \centering
  \includegraphics[scale=0.54]{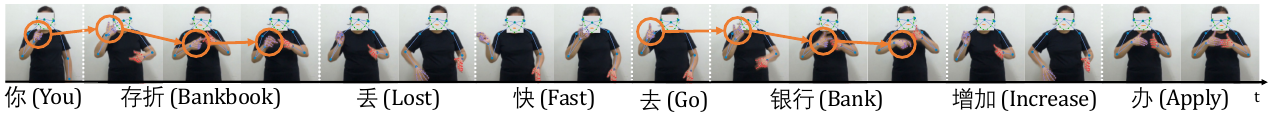}
  \caption{Visualization derived from the CSL-Daily dataset. The right hand, along with its primary motion trajectory, is highlighted to illustrate the gesture dynamics. The example corresponds to the sentence: \begin{CJK}{UTF8}{gbsn}“存折丢了的话要马上去银行补办。”\end{CJK}
 (If your bankbook is lost, you should go to the bank immediately to have it reissued.) This figure illustrates visually similar signs such as “bankbook” and “bank”, as well as “you” and “go”. Despite sharing highly similar motion patterns, each gesture serves a distinct syntactic and semantic function within the sentence. This example demonstrates the limitations of purely visual recognition and emphasizes the importance of strong visual-linguistic alignment for effective SLU.}
  \label{fig:vis_csl}
\end{figure}

We identified three key challenges faced by current SLP-based SLU methods in the Introduction of the main paper: weak semantic grounding, imbalanced local-global feature modelling, and ineffective cross-modal alignment. These issues frequently manifest in practical scenarios where visually similar gestures convey entirely different meanings depending on their context, temporal structure, or semantic function. Figures \ref{fig:vis_wlasl} and \ref{fig:vis_csl} provide visual examples drawn from the WLASL2000 and CSL-Daily datasets, respectively, illustrating how these challenges affect SLU. Note that the figures show selected frames for clarity, rather than the full sequence.

\subsection{Weak semantic grounding.} 
In the CSL-Daily example shown in Figure \ref{fig:vis_csl}, the sign sequence includes terms such as “bankbook” and “bank”, as well as “you” and “go”, which share similar hand shapes and spatial trajectories. Although these gestures appear visually alike, each one conveys a distinct meaning and serves a different syntactic function within the sentence. If a model focuses only on superficial motion or shape patterns without understanding the linguistic intent behind each gesture, it may generate inaccurate or overly generic translations. This example emphasizes the importance of semantic grounding, where models should recognize what is being signed and understand its meaning within the broader linguistic and contextual framework.

\subsection{Local-global imbalance.} 
The WLASL2000 examples shown in Figure \ref{fig:vis_wlasl} present two sign sequences, “Always” and “Someone,” which share highly similar hand shapes and motion trajectories across several frames. The primary distinction lies in the broader spatial and temporal extent of “Always” compared to the more confined gesture of “Someone.” Relying solely on local visual cues such as hand configuration or position is insufficient for accurate interpretation. At the same time, global cues alone cannot resolve subtle variations in form that are crucial for meaning. Accurate understanding requires the integration of fine-grained local details with the overarching motion pattern and semantic context. This example underscores the essential role of modelling both local and global features together. Only by combining detailed gesture recognition with a coherent understanding of the full temporal sequence can models distinguish between signs that are visually similar but semantically different.

\subsection{Ineffective cross-modal alignment.} 
Although Figures \ref{fig:vis_wlasl} and \ref{fig:vis_csl} highlight different challenges, both reveal a deeper problem rooted in weak alignment between visual and textual modalities. In Figure \ref{fig:vis_wlasl}, distinguishing between “Always” and “Someone” involves more than recognizing visual patterns. It requires establishing a clear connection between the motion sequence and its corresponding linguistic meaning. Similarly, in Figure \ref{fig:vis_csl}, the model should determine whether a gesture refers to “bank” or “bankbook,” even when the visual cues appear highly similar. Accurate interpretation depends on correctly linking each visual segment to its intended word or phrase within a broader sentence. Without a strong mechanism for aligning gestures with language, the model fails to generate consistent and meaningful outputs. These examples show that SLU is not just a visual recognition problem; it is a multimodal challenge that requires precise mapping from gestures to language at both lexical and semantic levels.

These visualizations serve as motivating evidence for the limitations (as discussed in the Introduction section of the main paper) of existing SLU approaches and the need for semantically informed modelling. Our proposed framework mitigates the impact of these problems by enriching visual features with linguistic context, balancing local and global feature interactions, and learning aligned cross-modal representations. 

\section{Advances in skeleton-based SLU}
A growing line of work explores skeleton as a compact and semantically informative modality for SLU. Early work such as GCN-BERT \citep{tunga2021pose} integrates graph convolutional networks over human joint graphs with transformer encoders, modelling spatial–temporal cues from skeletal sequences. Although effective for isolated SLR, it remains fully supervised and task-specific. To the best of our acknowledged, SignBERT \cite{hu2021signbert} is the first work applies self-supervised pre-training for hand-centric skeletal representations. By masking and reconstructing hand trajectories and leveraging an explicit hand-shape model for regularisation, it learns richer visual embeddings and improves both isolated and continuous recognition. Building on stronger structural modelling, BEST \citep{zhao2023best} advances skeleton-based pre-training by grouping body and hands into skeletal triplets and adopting a BERT-style masked unit modelling approach. A discrete VAE is used to tokenise continuous skeletal units into pseudo-tokens, enabling cross-entropy reconstruction and encouraging contextual reasoning over articulated hand–body interactions. BEST \citep{zhao2023best} demonstrates strong generalisation across isolated SLR benchmarks. SignBERT+ \cite{hu2023signbertplus} further extends this family of work by incorporating linguistic signals, and refining the pre-training tasks to support SLR and SLT. Compared with its predecessor, it provides more structured multi-task learning and better alignment between skeletal sequences and semantics. These methods reveal the strong potential of skeleton-only pre-training. However, they focus on capturing visual cues, lack joint modeling on the textual information, and remain visually grounded and largely task-specific. In contrast, Sigma builds on this foundation while moving toward a unified SLU paradigm and learns cross-modal alignment suitable for ISLR, CSLR, and SLT within a single framework.

\section{Cluster aggregator} \label{sec:cu_aggregator}
To address the challenges of weak semantic grounding and limited cross-modal alignment, we design a cluster aggregator module inspired by \citep{hou2024bagformer} to produce cluster-level textual embeddings that better correspond to visual sign units. Given a text input such as “curiosity,” the tokenizer splits it into subword tokens, which are then processed by the text encoder to generate token-level features. The aggregator groups these features into semantically coherent clusters. An offset calculator maps each original token to its cluster index, and the aggregation helper combines features within each cluster to form a compact representation. This process yields text embeddings that preserve semantic structure while reducing redundancy. Sigma supports both fine-grained gesture recognition and high-level translation. This mechanism contributes to more effective cross-modal learning and helps bridge the gap between dynamic visual input and structured language representations.

\section{Sign-grounded text encoder} \label{sec:sgt_enc}
To mitigate the impact of weak semantic grounding and ineffective cross-modal alignment, we enhance text representations by integrating visual cues from sign features through a dual-path architecture. This SGT encoder consists of two parallel branches: a sign-text matching (STM) path and a language modelling (LM) path. The STM path, repeated $M$ times, cooperates with cross-attention layers where textual tokens attend to sign features, allowing the model to align linguistic units with visual semantics and enrich textual embeddings with sign language gesture context. The LM path, repeated $N$ times, uses standard transformer blocks with self-attention and feed-forward layers to preserve language fluency and syntactic structure. This dual-path setup enables the SGT encoder to learn representations that are both semantically grounded in visual input and linguistically coherent. During fine-tuning, all parameters except for the self-attention layers within the\textbf{STM} path are transferred, ensuring effective knowledge reuse while allowing flexible adaptation to downstream SLU tasks. This design supports stronger cross-modal alignment and helps mitigate the semantic disconnect between dynamic sign inputs and static textual outputs.

\section{Skeletal data} \label{sec:why_skl_data}
\begin{figure}[htbp] 
  \centering
  \includegraphics[scale=0.35]{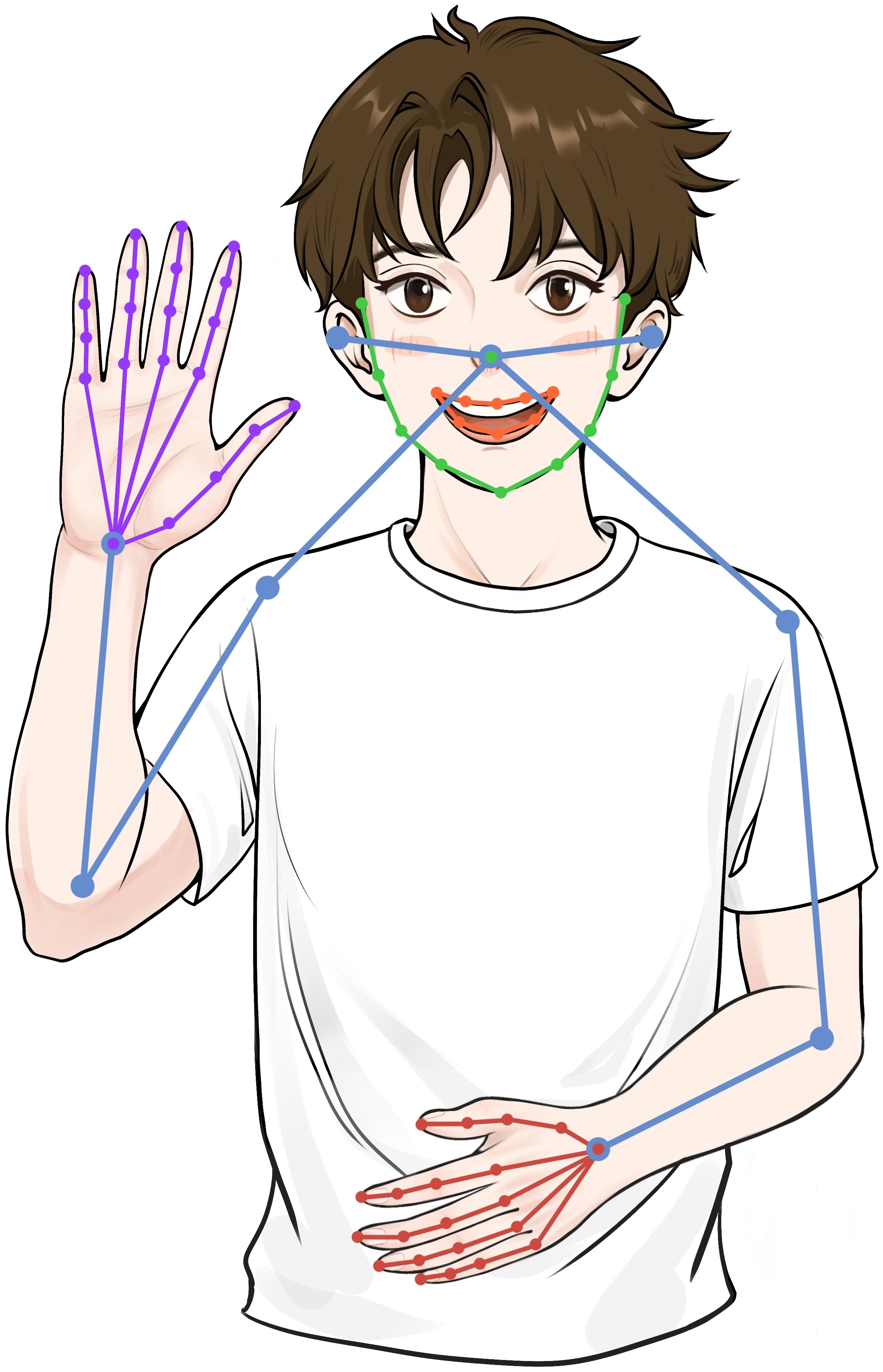}
  \caption{The visualisation of the full-body keypoints.}
  \label{fig:keypoint}
\end{figure}

The sign sequences are skeletal data extracted using RTMPose \citep{jiang2023rtmpose} from MMPose \citep{sengupta2020mm}. Figure \ref{fig:keypoint} illustrates the visualization of 69 keypoints per frame, including 21 for each hand, 9 for the body, and 18 for the face.

\section{What makes annotations costly in sign language processing?} \label{sec:cost_annottn}
In sign language processing, annotations refer to manually labelled data that describe the content and structure of sign language videos. These annotations are essential for training supervised learning models, but are significantly more expensive and labour-intensive than those in natural language processing.

There are three main reasons why annotations in this domain are costly:

1) \textbf{Expert-dependent labelling :} Unlike speech or text, sign language does not have a widely standardised written form. Annotators must label each gesture with its corresponding gloss, a textual representation of the meaning of the sign. This requires a deep level of linguistic expertise in both the sign language and the spoken language to which it is assigned. It is time-consuming, and the availability of such annotations is limited.

2) \textbf{Temporal segmentation and alignment:} For CSLR and SLT tasks, annotators must align glosses with precise time frames in sign language videos. Unlike tokenising text, this process requires identifying the exact start and end points of each sign within a continuous, unsegmented motion stream. Such fine-grained temporal labelling demands both visual precision and linguistic expertise, making the task exceptionally labour-intensive. In our study, temporal boundary labels are not used; glosses are only employed for ISLR and CSLR. With the growing availability of public sign language datasets, we hope that both ISLR and CSLR can eventually be learned without relying on any costly gloss annotations.

3) \textbf{Multi-layer multimodal cues:} sign language relies on hand gestures, facial expressions, body posture, and spatial references. Annotating these multimodal components accurately requires frame-by-frame observation and sometimes multi-camera viewpoints. Capturing this richness adds both time and complexity to the annotation process.

Due to these factors, building large-scale annotated datasets for SLU or sign language tasks remains a major bottleneck. This motivates the development of SLP-based SLU models as well as the use of self-supervised and weakly supervised methods, which can learn meaningful representations from unannotated or minimally annotated data.

\section{Rethinking the role of glosses in SLU} \label{sec:gloss_problem}
Gloss annotations have long been used as an intermediate representation in sign language translation, and they provide efficient and powerful supervision. By reducing the gap between raw visual input and spoken language output, glosses offer a structured, linguistically meaningful signal that has boosted SLT performance compared to purely end-to-end gloss-free approaches \citep{zhou2023gloss}. At the same time, this benefit comes with significant drawbacks that increasingly limit scalability and linguistic fidelity. First, glosses are costly to obtain and difficult to scale. Producing them requires expert annotators and fine-grained temporal alignment, making data collection expensive and slow, and constraining the size of available datasets. Second, glosses act as an information bottleneck. A gloss sequence compresses rich and continuous sign expressions into discrete tokens, discarding nuances not represented in the gloss inventory and weakening the direct mapping from visual input to textual semantics. Third, gloss-based pipelines suffer from error propagation and mismatched objectives. Since they typically rely on a continuous sign language recognition front end, recognition errors are passed into the translation stage, while training remains split across recognition and translation tasks rather than being optimised jointly. Moreover, glosses often fail to capture divergences between sign language structure and the linear order of spoken languages. By committing to a single gloss sequence early, models risk locking in alignment hypotheses that hinder later reordering and discourse modelling. Another limitation lies in the inability of glosses to encode non-manual signals such as facial expressions, or the compositional use of multiple articulators, both of which carry critical linguistic meaning. In addition, gloss conventions vary across datasets and languages, introducing inconsistencies that reflect annotation practices rather than genuine linguistic differences, which in turn hinders transfer learning and cross-corpus pre-training. Finally, reliance on gloss labels restricts data efficiency. Although gloss supervision can enhance performance when available, it blocks the use of large unlabelled video corpora, whereas direct visual–text modelling allows learning from broader resources. Taken together, these issues highlight that glosses provide strong but narrow supervision: while effective in guiding alignment, they remain a bottleneck for scalability. This motivates our use of cluster-wise contrastive learning, which produces gloss-like groupings automatically and retains the advantages of structured alignment while avoiding the limitations of manual gloss annotation.

\section{Complete results}

The complete SLT results for the OpenASL and CSL-Daily datasets, including all intermediate evaluation metrics across the DEV and TEST splits, are detailed in Table \ref{tab:openasl_result} and Table \ref{tab:csl_result}, respectively.

\begin{table}[htbp]
\centering
\caption{SLT results on OpenASL dataset.}
\resizebox{\textwidth}{!}{%
\begin{tabular}{lcccccccccc} 
\toprule
\multirow{2}{*}{Method} & \multicolumn{5}{c}{DEV} & \multicolumn{5}{c}{TEST}               \\ 
\cmidrule(r){2-6}\cmidrule{7-11}
                        & B@1$\uparrow$   & B@2$\uparrow$   & B@3$\uparrow$   & B@4$\uparrow$   & R@L$\uparrow$
                        & B@1$\uparrow$   & B@2$\uparrow$   & B@3$\uparrow$   & B@4$\uparrow$   & R@L$\uparrow$    \\ 
\cmidrule[\heavyrulewidth]{1-6}\cmidrule{7-11}

GloFE-VN \citep{lin2023gloss}                
& 21.06 & 12.34 & 8.68  & 6.68  & 21.37                  
& 21.56 & 12.74 & 9.05  & 7.06  & 21.75  \\

Conv-GRU \citep{camgoz2018neural}                
& 16.72 & 8.95  & 6.31  & 4.82  & 16.25                  
& 16.11 & 8.85  & 6.18  & 4.58  & 16.10  \\

I3D-transformer \citep{shi2022open}         
& 18.26 & 10.26 & 7.17  & 5.60  & 18.88                  
& 18.31 & 10.15 & 7.19  & 5.56  & 18.64  \\

OpenASL \citep{shi2022open}                 
& 20.10 & 11.81 & 8.43  & 6.57  & 20.43                  
& 20.92 & 12.08 & 8.59  & 6.72  & 21.02  \\

Uni-Sign \citep{Li2025sign}               
& 50.84 & 37.82 & 29.83 & 24.16 & 44.58                  
& 49.35 & 36.32 & 28.55 & 23.14 & 43.22  \\

$C^{2}RL$ \citep{chen2025c}                    
& -     & -     & -     & -     & -                      
& 31.46 & 21.85 & 16.58 & 13.21 & 31.36  \\

\textbf{Sigma} 
& \textbf{51.68}
& \textbf{37.90}
& \textbf{30.35}
& \textbf{25.72}
& \textbf{45.81}
& \textbf{49.91}
& \textbf{36.89}
& \textbf{28.80}
& \textbf{23.21}
& \textbf{45.38} \\

\bottomrule
\end{tabular}%
}
\label{tab:openasl_result}
\end{table}

\begin{table}[htbp]
\centering
\caption{SLT results on CSL-Daily dataset.}
\resizebox{\textwidth}{!}{%
\begin{tabular}{clcccccccccc} 
\toprule
 & Method 
& \multicolumn{5}{c}{DEV}
& \multicolumn{5}{c}{TEST} \\
\cmidrule(lr){3-7}\cmidrule(lr){8-12}
& 
& B@1$\uparrow$ & B@2$\uparrow$ & B@3$\uparrow$ & B@4$\uparrow$ & R@L$\uparrow$
& B@1$\uparrow$ & B@2$\uparrow$ & B@3$\uparrow$ & B@4$\uparrow$ & R@L$\uparrow$ \\
\midrule

\multirow{7}{*}{\rotatebox{90}{Gloss-based}}
& SLRT \citep{camgoz2020sign}
& 37.47 & 24.67 & 16.86 & 11.88 & 37.96
& 37.38 & 24.36 & 16.55 & 11.79 & 36.74 \\


& SignBT \citep{zhou2021improving}
& 51.46 & 37.23 & 27.51 & 20.80 & 49.49
& 51.42 & 37.26 & 27.76 & 21.34 & 49.31 \\

& MMTLB \citep{chen2022simple}
& 53.81 & 40.84 & 31.29 & 24.42 & 53.38
& 53.31 & 40.41 & 30.87 & 23.92 & 53.25 \\

& SLTUNET \citep{zhang2023sltunet}
& - & - & - & 23.99 & 53.58
& 54.98 & 41.44 & 31.84 & 25.01 & 54.08 \\

& TS-SLT \citep{chen2022two}
& 55.21 & 42.31 & 32.71 & 25.76 & 55.10
& 55.44 & 42.59 & 32.87 & 25.79 & 55.72 \\

& CV-SLT \citep{zhao2024conditional}
& 58.05 & 44.73 & 35.14 & 28.24 & 56.36
& 58.29 & 45.15 & 35.77 & 28.94 & 57.06 \\

\midrule

\multirow{11}{*}{\rotatebox{90}{Gloss-free}}
& SLRT \citep{camgoz2020sign}
& 21.03 & 9.97 & 5.96 & 4.04 & 20.51
& 20.00 & 9.11 & 4.93 & 3.03 & 19.67 \\

& GASLT \citep{yin2023gloss}
& - & - & - & - & -
& 19.90 & 9.94 & 5.98 & 4.07 & 20.35 \\

& MSLU \citep{zhou2024scaling}
& 33.28 & 21.31 & - & 10.27 & 33.13
& 33.97 & 22.20 & - & 11.42 & 33.80 \\

& NSLT \citep{camgoz2018neural}
& 34.22 & 19.72 & 12.24 & 7.96 & 34.28
& 34.16 & 19.57 & 11.84 & 7.56 & 34.54 \\

& GFSLT-VLP \citep{zhou2023gloss}
& 39.20 & 25.02 & 16.35 & 11.07 & 36.70
& 39.37 & 24.93 & 16.26 & 11.00 & 36.44 \\

& FLa-LLM \citep{chen2024factorized}
& - & - & - & - & -
& 37.13 & 25.12 & 18.38 & 14.20 & 37.25 \\

& $C^{2}RL$ \citep{chen2025c}
& - & - & - & - & -
& 49.32 & 36.28 & 27.54 & 21.61 & 48.21 \\

& Uni-Sign \citep{Li2025sign}
& 55.30 & 42.21 & 32.94 & 26.25 & 56.03
& 55.08 & 42.14 & 32.98 & 26.36 & 56.51 \\

& SignLLM \citep{gong2024llms}
& 42.45 & 26.88 & 17.90 & 12.23 & 39.18
& 39.55 & 28.13 & 20.07 & 15.75 & 39.91 \\

& Sign2GPT \citep{wong2024sign2gpt}
& - & - & - & - & -
& 41.75 & 28.73 & 20.60 & 15.40 & 42.36 \\

& \textbf{Sigma}
& \textbf{56.12} & \textbf{47.12} & \textbf{34.17} & \textbf{27.80} & \textbf{57.12}
& \textbf{55.73} & \textbf{43.42} & \textbf{33.12} & \textbf{27.12} & \textbf{57.32} \\

\bottomrule
\end{tabular}%
}
\label{tab:csl_result}
\end{table}

\section{Optimising local cluster-wise contrastive learning strategies} \label{sec:cw-cl_design_choice}
Local cluster-wise contrastive learning plays a key role in capturing fine-grained visual-text correspondence. Unlike global alignment that operates on sentence-level representations, local alignment focuses on token-level interactions between sign features and textual clusters. This design enables the model to discover semantically meaningful correspondences between continuous sign motions and clustered textual units that approximate gloss-like structures.

The overall computation pipeline is illustrated in Figure~\ref{fig:local-cw-sim}. First, the similarity matrix $M$ is constructed between sign features and textual clusters. Next, a row-wise max operation extracts the strongest alignment signal for each visual token to get the row max matrix $R$. The resulting scores are then normalised using a softmax function to produce token weights, which are used to compute weighted similarity $scores$. Finally, in-batch local contrastive learning is applied to maximize the similarity $\mathbf{M}^{l}_{\mathrm{s2t}}$ of matched sign–text pairs (filled by light orange) while minimising that of mismatched pairs.

This design encourages the model to focus on the most semantically relevant cross-modal correspondences while maintaining robust aggregation of token-level information. As a result, the proposed local cluster-wise contrastive learning mechanism provides stronger fine-grained alignment between visual sign segments and clustered textual representations, thereby improving semantic grounding in sign language understanding.

\begin{figure}[htbp]
  \centering
  \includegraphics[scale=0.58]{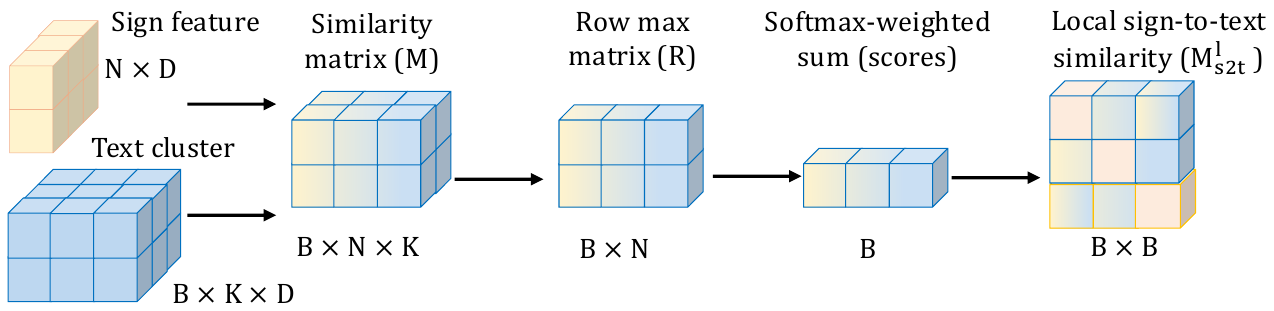}
  \caption{Illustration of the computation of our local sign-to-text cluster-wise similarity inspired by \citep{chen2020simple, radford2021learning, li2022blip, hou2024bagformer}. The similarity matrix $M$ is computed between the sign feature of each sample and all textual clusters. For each sample, the maximum similarity score is computed using a max operation, which forms $R$. The resulting values are passed through a softmax-weighted sum function to obtain the local similarity $scores$. Finally, in-batch local cluster contrastive learning is applied to pull semantically aligned visual-text pairs (highlighted in light orange) closer together, while pushing apart unaligned pairs. This process enables localised semantic grounding by focusing on the most relevant visual-text associations within each cluster.}
  \label{fig:local-cw-sim}
\end{figure}

\section{Additional experiment} \label{sec:ad_ablation}
In this section, we provide additional ablation studies and extended comparisons to further analyse the behaviour of the proposed model under different design choices. Unless otherwise specified, all experimental settings follow those in the main paper. Best results in \colorbox{darkb}{dark blue} and second best in \colorbox{lightb}{light blue}.

\subsubsection{Row-wise operations} 
To compute cluster-wise similarity (as illustrated in Figure \ref{fig:local-cw-sim}) in an optimal way, we evaluate several row-wise operations in Table \ref{tab:row_operations}, including row max, average, top-$k$ average, and softmax-based operation. For the top $k$ average, the value of $k$ is dynamically determined based on the number of clusters or tokens that exist in the cosine similarity matrix $M$, using the formula: 

$ k = \max \left( 1, \left\lfloor \frac{M}{3} \right\rfloor \right) $

This ensures that $k$ remains a valid positive integer bounded by the length of the last dimension of $M$, with a lower bound of 1 to avoid degenerate cases. In addition to max and average operation over the similarity matrix $M$, we also evaluated a softmax-based operation expressed as:

$
R = \mathrm{sum}(\mathrm{softmax}(M) \odot M,\; dim=1)
$

Empirical results in Table~\ref{tab:row_operations} demonstrate that the row-wise max operation consistently delivers the best overall performance across ISLR, CSLR, and SLT tasks.

Specifically, on CSLR (CSL-Daily), row max achieves the lowest WER of 25.26, outperforming row average (25.40), row top-$k$ average (25.88), and row softmax (25.50). A similar trend is observed for SLT on CSL-Daily, where row max attains the highest BLEU-4 score of 27.12 and the best ROUGE-L of 57.32, exceeding row average (26.78 / 56.60), top-$k$ average (26.48 / 57.08), and softmax (26.85 / 56.75). On ISLR (WLASL2000), row max also achieves the strongest performance with 64.54 P-I and 62.28 P-C accuracy.

Although alternative aggregation strategies such as top-$k$ averaging and softmax are designed to smooth cluster-level similarities and reduce noise, the empirical results suggest that such smoothing may attenuate the most discriminative alignment signals. In contrast, the max operator directly emphasises the strongest cluster-level correspondence, yielding sharper contrastive gradients and more decisive cross-modal alignment.

These findings indicate that preserving the most salient matching signal is more beneficial than distributing weights across multiple clusters. Consequently, we adopt row-wise max as the default aggregation strategy in our local cluster-wise contrastive learning.

\begin{table}[htbp]
  \centering
  \caption{Row operations for local cluster-wise contrastive learning.}
  \label{tab:row_operations}
  \resizebox{0.8\textwidth}{!}{%
  \begin{tabular}{lcccccc} 
  \toprule
  \multirow{3}{*}{Strategy} & \multicolumn{2}{c}{WLASL2000} & \multicolumn{4}{c}{CSL-Daily} \\ 
  \cmidrule(lr){2-3} \cmidrule(lr){4-7}
                            & \multicolumn{2}{c}{ISLR}      & CSLR  & \multicolumn{3}{c}{SLT} \\
                            & P-I↑  & P-C↑                   & WER↓  & B@1↑  & B@4↑  & R@L↑ \\
  \midrule

  Row max 
  & \cellcolor{darkb}\textbf{64.54}
  & \cellcolor{darkb}\textbf{62.28}
  & \cellcolor{darkb}\textbf{25.26}
  & \cellcolor{darkb}\textbf{55.73}
  & \cellcolor{darkb}\textbf{27.12}
  & \cellcolor{darkb}\textbf{57.32} \\

  Row average
  & 64.18
  & \cellcolor{lightb}62.123
  & \cellcolor{lightb}25.40
  & 55.123
  & 26.78
  & 56.60 \\

  Row top-$k$ average
  & \cellcolor{lightb}64.32
  & 62.02
  & 25.88
  & \cellcolor{lightb}55.48
  & 26.48
  & \cellcolor{lightb}57.08 \\

  Row softmax
  & 64.10
  & 62.05
  & 25.50
  & 55.22
  & \cellcolor{lightb}26.85
  & 56.75 \\

  \bottomrule
  \end{tabular}}
\end{table}

\subsubsection{Local-level scoring} 
To evaluate different local-level scoring strategies for cluster-wise contrastive learning, we compare several accumulation methods in Table \ref{tab:local_scoring}. These include basic aggregation approaches (sum and average) as well as more expressive formulations such as log-sum-exp, softmax, and variance-reduced-sum.

Across the benchmarks listed in Table \ref{tab:local_scoring_results}, the softmax scoring method consistently achieves the best overall performance. On CSLR (CSL-Daily), softmax yields the lowest WER of 25.26, outperforming sum (25.88), average (26.30), log-sum-exp (25.40), and variance-reduced-sum (25.60). A similar pattern is observed for SLT, where softmax achieves the highest BLEU-4 (27.12) and ROUGE-L (57.32), exceeding all alternative strategies. On ISLR (WLASL2000), softmax also achieves the strongest recognition accuracy with 64.54 P-I and 62.28 P-C.

While log-sum-exp and variance-reduced-sum provide competitive results by offering smoother approximations of maximum similarity, their performance remains slightly inferior and less consistent across tasks. In contrast, softmax dynamically assigns probabilistic weights to token-level similarities, enabling the model to emphasise highly informative alignments while retaining contextual diversity. 

These findings suggest that adaptive weighting is more effective than uniform aggregation or hard summation when modelling fine-grained cross-modal correspondences. Therefore, we adopt softmax as the default local-level scoring function in our cluster-wise contrastive learning framework.

\begin{table}[htbp]
\centering
\caption{Local-level scoring methods.}
\label{tab:local_scoring}
\resizebox{0.8\textwidth}{!}{%
\begin{tabular}{ll}
\toprule
\textbf{Scoring Method} & \textbf{Pseudocode} \\ 
\midrule
Softmax & $score \leftarrow \mathrm{sum}(\mathrm{softmax}(R) \odot R,\; dim=1)$ \\
Log-sum-exp & $score \leftarrow \log(\mathrm{sum}(\exp(R),\; dim=1))$ \\
Variance-reduced-sum & $score \leftarrow \mathrm{sum}((R - \mathrm{mean}(R,\; dim=1)),\; dim=1)$ \\
\bottomrule
\end{tabular}%
}
\end{table}

\begin{table}[htbp]
  \centering
  \caption{Local-level scoring methods for the local cluster-wise contrastive learning.}
  \label{tab:local_scoring_results}
  \resizebox{0.8\textwidth}{!}{%
  \begin{tabular}{lcccccc} 
  \toprule
  \multirow{3}{*}{Strategy} & \multicolumn{2}{c}{WLASL2000} & \multicolumn{4}{c}{CSL-Daily} \\ 
  \cmidrule(lr){2-3} \cmidrule(lr){4-7}
                            & \multicolumn{2}{c}{ISLR}      & CSLR  & \multicolumn{3}{c}{SLT} \\
                            & P-I↑  & P-C↑                   & WER↓  & B@1↑  & B@4↑  & R@L↑ \\
  \midrule

  Sum
  & 64.32
  & 62.05
  & 25.88
  & 55.28
  & 26.94
  & 56.95 \\

  Average
  & 64.20
  & 61.92
  & 26.30
  & 55.10
  & 26.70
  & 56.70 \\

  Log-sum-exp
  & 64.38
  & \cellcolor{lightb}62.22
  & \cellcolor{lightb}25.40
  & \cellcolor{lightb}55.55
  & \cellcolor{lightb}27.02
  & \cellcolor{lightb}57.18 \\

  Softmax
  & \cellcolor{darkb}\textbf{64.54}
  & \cellcolor{darkb}\textbf{62.28}
  & \cellcolor{darkb}\textbf{25.26}
  & \cellcolor{darkb}\textbf{55.73}
  & \cellcolor{darkb}\textbf{27.12}
  & \cellcolor{darkb}\textbf{57.32} \\

  Variance-reduced-sum
  & \cellcolor{lightb}64.46
  & 62.10
  & 25.60
  & 55.40
  & 26.88
  & 57.05 \\

  \bottomrule
  \end{tabular}}
\end{table}

\subsection{Should the Text Encoder Be Trainable During Pre-training?}

We further investigate whether the text encoder should remain frozen or be updated during pre-training. The comparison is reported in Table~\ref{tab:unfreeze?}. 

Empirically, allowing the text encoder to be trainable consistently yields better performance across all tasks. On CSL-Daily, unfreezing the text encoder reduces CSLR WER from 25.68 to 25.26. For SLT, it improves BLEU-4 from 26.90 to 27.12 and ROUGE-L from 56.95 to 57.32. Similar gains are observed on ISLR (WLASL2000), where P-I accuracy increases from 64.40 to 64.54 and P-C from 62.14 to 62.28.

Although the absolute improvements are moderate, the trend is consistent across recognition and translation benchmarks. This indicates that jointly optimising the text encoder during pre-training facilitates better cross-modal alignment. By adapting textual representations to skeleton-based visual features, the model can establish more coherent semantic correspondences, which subsequently benefit downstream tasks.

These results suggest that freezing the text encoder may constrain representation alignment, whereas end-to-end optimisation enhances cross-modal compatibility. Therefore, we adopt the unfreezed setting as the default configuration in Sigma.

\begin{table}[htbp]
  \centering
  \caption{Impact of freezing text encoder during pre-training.}
  \label{tab:unfreeze?}
  \resizebox{0.75\textwidth}{!}{%
  \begin{tabular}{lcccccc} 
  \toprule
  \multirow{3}{*}{Strategy} & \multicolumn{2}{c}{WLASL2000} & \multicolumn{4}{c}{CSL-Daily} \\ 
  \cmidrule(lr){2-3} \cmidrule(lr){4-7}
                            & \multicolumn{2}{c}{ISLR}      & CSLR  & \multicolumn{3}{c}{SLT} \\
                            & P-I↑  & P-C↑                   & WER↓  & B@1↑  & B@4↑  & R@L↑ \\
  \midrule

  Freezed
  & \cellcolor{lightb}64.40
  & \cellcolor{lightb}62.14
  & \cellcolor{lightb}25.68
  & \cellcolor{lightb}55.38
  & \cellcolor{lightb}26.90
  & \cellcolor{lightb}56.95 \\

  Unfreezed
  & \cellcolor{darkb}\textbf{64.54}
  & \cellcolor{darkb}\textbf{62.28}
  & \cellcolor{darkb}\textbf{25.26}
  & \cellcolor{darkb}\textbf{55.73}
  & \cellcolor{darkb}\textbf{27.12}
  & \cellcolor{darkb}\textbf{57.32} \\

  \bottomrule
  \end{tabular}}
\end{table}

\begin{table}[htbp]
  \centering
  \caption{Impact of different feature modalities.}
  \label{tab:df_features}
  \resizebox{0.75\textwidth}{!}{%
  \begin{tabular}{lcccccc} 
  \toprule
  \multirow{3}{*}{Strategy} & \multicolumn{2}{c}{WLASL2000} & \multicolumn{4}{c}{CSL-Daily} \\ 
  \cmidrule(lr){2-3} \cmidrule(lr){4-7}
                            & \multicolumn{2}{c}{ISLR}      & CSLR  & \multicolumn{3}{c}{SLT} \\
                            & P-I↑  & P-C↑                   & WER↓  & B@1↑  & B@4↑  & R@L↑ \\
  \midrule

  Sign feature
  & \cellcolor{darkb}\textbf{64.54}
  & \cellcolor{darkb}\textbf{62.28}
  & \cellcolor{darkb}\textbf{25.26}
  & \cellcolor{darkb}\textbf{55.73}
  & \cellcolor{darkb}\textbf{27.12}
  & \cellcolor{darkb}\textbf{57.32} \\

  Text feature
  & \cellcolor{lightb}64.41
  & \cellcolor{lightb}62.16
  & \cellcolor{lightb}25.34
  & \cellcolor{lightb}55.65
  & \cellcolor{lightb}27.05
  & \cellcolor{lightb}57.18 \\

  \bottomrule
  \end{tabular}}
\end{table}

\subsection{Cooperation of Different Modalities During Pre-training}
To examine how different feature modalities contribute to cross-modal pre-training, we compare using sign features and text features as inputs to the cross-attention module in the SGT encoder, as shown in Table~\ref{tab:df_features}.

Both modalities independently support strong downstream performance, indicating that our pre-training framework effectively extracts transferable semantic representations from either visual or textual sources. However, using sign features consistently achieves slightly better results across all benchmarks. On CSLR (CSL-Daily), sign features reduce WER from 25.34 to 25.26. For SLT, sign features improve BLEU-4 from 27.05 to 27.12 and ROUGE-L from 57.18 to 57.32. Similarly, on ISLR (WLASL2000), sign features achieve higher P-I (64.54 vs. 64.41) and P-C accuracy (62.28 vs. 62.16).

Although the performance gap is relatively small, the trend is consistent. This suggests that directly modelling fine-grained visual dynamics during pre-training provides slightly stronger supervision signals than relying on textual representations alone. Text features remain effective due to their structured linguistic information, but sign features better preserve motion-level discriminative cues.

Overall, the results confirm that both modalities are informative for semantic alignment, while visual sign features offer marginally stronger cross-modal grounding. This validates the flexibility of our framework and the robustness of the proposed cross-modal pre-training strategy.

\section{Ethics Statement}
Our work focuses on sign language understanding, aiming to improve accessibility and communication for people with hearing or speech impairment. The datasets we use (WLASL, CSL-Daily, How2Sign, OpenASL) are publicly available and widely adopted in SLU research. We strictly follow dataset licenses and use them only for academic purposes. No personally identifiable information or sensitive attributes beyond the original releases are introduced. We acknowledge the cultural and linguistic importance of sign languages and stress that our models are intended to support accessibility rather than replace human interpreters.

\section{Reproducibility Statement}
We ensure reproducibility by providing detailed descriptions of our models, objectives, and training configurations in the main text and appendix. Dataset statistics and preprocessing steps are clearly reported. Hyperparameters, loss formulations, and evaluation protocols are included to enable replication. To further support reproducibility, we will release our code upon publication. These resources will allow the community to replicate our experiments and extend our work on sign language understanding.

\bibliographystyle{elsarticle-num} 
\bibliography{main}

\end{document}